\documentclass{article} 
\usepackage{iclr2017_conference,times}
\usepackage{hyperref}
\usepackage{url}
\usepackage{enumitem}
\usepackage{float}

\usepackage{amsthm}
\usepackage{amsmath}
\usepackage{amssymb}
\usepackage{color}
\usepackage{xspace}
\usepackage{pgfplots, pgfplotstable}

\newcommand{\yrcite}[1]{\citeyearpar{#1}}

\newcommand{\mohammad}[1]{\ifdefined\DRAFT{\textcolor{blue}{(Mo: #1)}}}

\newcommand{\todo}[1]{\ifdefined\DRAFT{\textcolor{brown}{\bf TODO: #1}}\else{}\fi}
\newcommand{\comment}[1]{}
\newcommand{\extracite}[1]{#1}

\def\deriv{\mathrm{d}}

\renewcommand{\vec}[1]{\boldsymbol{\mathbf{#1}}}

\def\init{g}
\def\transition{f}
\def\baseline{b}

\def\rlrml{UREX\xspace}
\def\rml{\text{UREX}\xspace}
\def\pg{\text{MENT}\xspace}

\def\expected{\mathbb{E}}

\def\bsmax{2n\!+\!1}
\def\bold{}

\def\btheta{\vec{\theta}}
\def\bphi{\vec{h}}
\def\bphii{\vec{h}^{(n)}}
\def\ba{\vec{a}}
\def\bat{\vec{a}}

\def\bast{\vec{a}^{(k)}}

\def\bs{\vec{s}}

\def\calY{\mathcal{Y}}

\def\modelp{{\pi}_{\theta}} 
\def\optp{{\pi}^*_\temp} 
\def\optpt{{\pi}^{*T}_\temp} 
\def\softmax{\mathrm{softmax}}

\def\imw{w_\temp}

\def\bys{{\mathbf y}^*}
\def\byp{{\mathbf y}} 

\def\temp{\tau}

\def\A{\mathcal{A}}

\def\objrml{\mathcal{O}_{\mathrm{RAML}}}
\def\objrlrml{\mathcal{O}_{\mathrm{\rlrml}}}

\def\objrl{\mathcal{J}_{\mathrm{RL}}}
\def\objrl{\mathcal{O}_{\mathrm{RL}}}
\def\objrwr{\mathcal{O}_{\mathrm{RWR}}}

\def\int{\mathrm{int}}

\newcommand{\reward}[2]{r(#1 \mid #2)}
\newcommand{\hatrewardd}[0]{\widehat{r}}
\newcommand{\hatreward}[2]{\widehat{r}\,(#1 \mid #2)}

\newcommand{\kl}[2]{D_{\mathrm{KL}}\left(#1~\Vert~#2\right)}

\newcommand{\ent}[1]{\mathbb{H}\left(#1\right)}

\def\eg{{\em e.g.,}}
\def\ie{{\em i.e.,}}
\def\iid{{\em i.i.d.}\xspace}

\def\vs{{\em vs.}\xspace}

\newcommand{\tabref}[1]{Table~\ref{#1}}
\newcommand{\figref}[1]{Figure~\ref{#1}}

\title{Improving Policy Gradient by\\Exploring Under-appreciated Rewards}

\author{
Ofir Nachum\thanks{Work done as a member of the Google Brain Residency program (\url{g.co/brainresidency})}~,
~Mohammad Norouzi,
~Dale Schuurmans\thanks{Also at the Department of Computing Science, University of Alberta, {\tt daes@ualberta.ca}}\\
~Google Brain\\
~\texttt{\{ofirnachum,\,mnorouzi,\,schuurmans\}@google.com}
}

\iclrfinalcopy 

\begin{document}

\maketitle


\vspace*{.5cm}
\begin{abstract}

This paper presents a novel form of policy gradient for model-free
reinforcement learning (RL) with improved exploration properties.
Current policy-based methods
use entropy regularization to encourage undirected exploration of the
reward landscape, which is ineffective in high dimensional spaces with
sparse rewards. We propose a more directed exploration strategy that
promotes exploration of {\em under-appreciated reward} regions.  An
action sequence is considered under-appreciated if its log-probability
under the current policy under-estimates its \mbox{resulting} reward.
The proposed exploration strategy is easy to implement, requiring
small modifications to 
the REINFORCE algorithm.
We evaluate the approach on a set of algorithmic tasks that have long
challenged RL methods. Our approach reduces hyper-parameter
sensitivity and demonstrates significant improvements over baseline
methods.  
The proposed
algorithm successfully solves a benchmark multi-digit
addition task and generalizes to long sequences, 
which,
to our
knowledge, 
is
the first time that a pure RL method has solved addition
using only reward feedback.

\comment{
This paper presents a novel form of policy gradient for model-free
reinforcement learning with improved exploration
characteristics. Current policy gradient methods use entropy
regularization to encourage uniform exploration of the reward
landscape, which is not effective in high dimensional spaces with
sparse rewards. We propose a better exploration strategy that
promotes exploration of the {\em under-appreciated reward} regions.
An action sequence is considered under-appreciated if its
\mbox{probability} under the current policy under-estimates its
resulting reward.  The proposed exploration strategy is easy to
implement. It improves robustness to the change of hyper-parameters,
and on a set of algorithmic tasks, it demonstrates significant
improvements against the policy gradient baselines. Our approach is
able to solve a benchmark multi-digit addition task.  To our
knowledge, this is the first pure RL method to solve addition
just by using reward feedback.
}

\comment{
To our knowledge Unlike standard on-line
learning strategies in reinforcement learning, which generally
encourage undirected exploration, we stochastically optimize a
modified objective that favors exploration of high-reward versus
low-reward actions.  A key feature of the algorithm is its ease of
application to recurrent neural network policy representations.  We
conduct an experimental evaluation on a set of algorithm induction
tasks, ranging from simple to challenging, and compare the proposed
enhancement to standard baselines.  In addition to obtaining notable
improvements, we also observe that the new approach can solve the
benchmark addition problem, which to our knowledge is the first
successful such demonstration based solely on success/failure
reinforcement.
}
\end{abstract}

\vspace*{.5cm}
\section{Introduction}

Humans can reason about symbolic objects and solve algorithmic problems.
After learning to count and then manipulate numbers via
simple arithmetic,
people eventually learn to invent new algorithms and even reason
about their correctness and efficiency.
The ability to invent new algorithms is fundamental to artificial
intelligence (AI).
Although symbolic reasoning has a long history in AI~\citep{AIbook}, 
only recently have statistical machine learning and neural network approaches begun to make 
headway in automated algorithm discovery
\citep{ReedF15, KaiserS15, NeelakantanLS15},
which would constitute
an important milestone on the path to AI.
Nevertheless, most of the recent successes depend on the use of strong supervision to
learn a mapping from a set of training inputs to outputs by maximizing a conditional
log-likelihood,
very much like neural machine translation systems
\citep{Sutskever14, bahdanau2014neural}. 
Such a dependence on strong supervision is a significant limitation
that does not match the ability of people to invent new algorithmic
procedures based solely on trial and error.

By contrast, {\em reinforcement learning (RL)} methods \citep{suttonbook}
hold the promise of searching over discrete objects such as symbolic representations of algorithms
by considering much weaker feedback in the form of a simple verifier that tests 
the correctness of a program execution on a given problem instance.
Despite the recent excitement around the use of RL to tackle
Atari games~\citep{atarinature} and Go~\citep{silveretal16},
standard RL methods are not yet able to consistently and reliably solve 
algorithmic tasks in all but the simplest cases~\citep{ZarembaS14}. 
A key property of algorithmic problems that makes them 
challenging for RL is {\em reward sparsity},
\ie~a policy usually has to get a long action sequence exactly right 
to obtain a non-zero reward.


We believe one of the key factors limiting the effectiveness of
current RL methods in a sparse reward setting is the use of {\em
undirected exploration} strategies~\citep{thrun92}, such
as \mbox{$\epsilon$-greedy} and entropy
regularization~\citep{williams1991function}.  For long action
sequences with delayed sparse reward, it is hopeless to explore the
space uniformly and blindly.  Instead, we propose a formulation to
encourage exploration of action sequences that are {\em
under-appreciated} by the current policy.  Our formulation considers
an action sequence to be under-appreciated if the model's
log-probability assigned to an action sequence under-estimates the
resulting reward from the action sequence.  Exploring
under-appreciated states and actions encourages the policy to have a
better calibration between its log-probabilities and observed reward
values, even for action sequences with negligible rewards. This
effectively increases exploration around neglected action sequences.

We term our proposed technique
{\em under-appreciated reward exploration (\rlrml)}. 
We show that the objective given by \rml is a combination of a
mode seeking objective (standard REINFORCE)
and a mean seeking term, which provides a
well motivated
trade-off between exploitation and exploration.
To empirically evaluate our method, we take a set of algorithmic tasks
such as sequence reversal, multi-digit addition, and binary search.  
We choose to focus on these tasks because, although simple, they present a
difficult sparse reward setting which has limited the success of standard RL approaches.
%
The experiments demonstrate that \rlrml significantly outperforms
baseline RL methods, such as entropy regularized REINFORCE and
one-step Q-learning, especially on the more difficult tasks, such as
multi-digit addition.  Moreover, \rlrml is shown to be more robust to
changes of hyper-parameters, which makes hyper-parameter tuning less
tedious in practice.  In addition to introducing a new variant of
policy gradient with improved performance, our paper is the first to
demonstrate strong results for an RL method on algorithmic tasks.  To
our knowledge, the addition task has not been solved by any model-free
reinforcement learning approach.  We observe that some of the policies
learned by \rlrml can successfully generalize to long sequences;
\eg~in $2$ out of $5$ random restarts, the policy learned by \rlrml
for the addition task correctly generalizes to addition of 
numbers with $2000$ digits with no mistakes, even though training
sequences are at most $33$ digits long.

\comment{
In order to find both non-zero reward action sequences and maximal
reward action sequences in a large action space, some sort of
exploration is necessary.  In value-based RL this exploration can
manifest as $\epsilon$-greedy actions whereas in policy-based methods
a common choice is to add an entropy
regularizer~\citep{williams1991function\extracite{,mnih2016asynchronous}}
to the well-known REINFORCE policy gradient
algorithm~\citep{Williams92}..  However, these exploration strategies
apply exploration blindly and uniformly.  In large action spaces,
these strategies can easily distract an agent from its main task of
honing in on high-reward regions of the policy space.

In this work, we propose a novel addition to the REINFORCE objective to favor
exploration in high-reward areas over exploration in low-reward areas.

We begin in Section \ref{related_work}, where we provide a short
survey of previous work on learning algorithms.  In Section
\ref{background} we provide background information regarding our
notation and the derivation of REINFORCE and its entropy-regularized
variant.  We introduce our approach and the resulting policy gradient
training algorithm in \ref{methodology}.  Subsequently, we present
several standard tasks and introduce an additional task we designed
ourselves in Section \ref{experimental_setup}, on which we then
empirically evaluate the performance of our approach.  We conclude in
Section \ref{conclusion}, where we summarize the paper and discuss the
repercussions and impact we hope it will have on future work.
}

\section{Neural Networks for Learning Algorithms}
\label{neural_algo}

Although research on using neural networks to learn algorithms has 
witnessed
a
surge of recent interest, the problem of program induction from
examples has a long history in many fields, including program
induction, inductive logic programming \citep{ilpbook}, relational
learning~\citep{kempetal07} and regular language
learning~\citep{angulin87}. Rather than presenting a comprehensive
survey of program induction here, we focus on neural network approaches to
algorithmic tasks and highlight the relative simplicity of our neural
network architecture.

Most successful applications of neural networks to algorithmic
tasks rely on strong supervision, where the inputs and target outputs
are completely known {\em a priori}. Given a dataset of examples, one
learns the network parameters by maximizing the conditional likelihood
of the outputs via backpropagation (\eg~\citet{ReedF15, KaiserS15,
  pointer_networks}). However, target outputs may not be available for
novel tasks, for which no 
prior algorithm is known to be available.
A more desirable
approach to inducing algorithms, followed in this paper, advocates
using self-driven learning strategies that only receive reinforcement
based on the outputs produced. Hence, just by having access to a
verifier for an algorithmic problem, one can aim to learn an
algorithm. For example, if one does not know how to sort an array, but
can check the extent to which an array is sorted, then one can provide
the reward signal necessary for learning sorting algorithms.

\comment{
However, providing the correct output during
training can often be infeasible; for example, when trying to solve a
problem with no known solution, or when efficiency is desired in
addition to correctness.
}

\begin{figure}
\begin{center}
\includegraphics[width=0.70\textwidth]{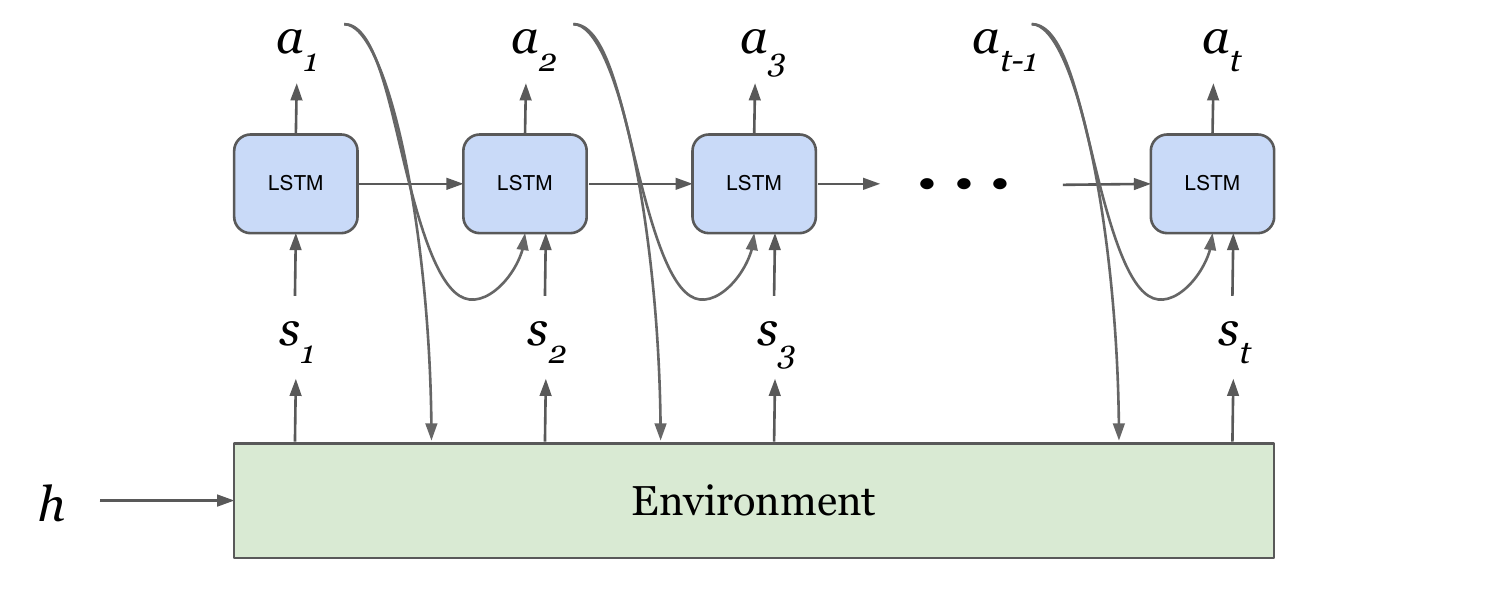}
\end{center}
\caption{
The agent's RNN architecture that represents a policy. The environment
is initialized with a latent vector $\bphi$. At time step $t$, the
environment produces a state $\bs_t$, and the agent takes as input
$\bs_t$ and the previously sampled action $\ba_{t-1}$ and produces a
distribution over the next action $\modelp(\ba_{t}\mid\bs_t)$. Then,
we sample a new action $\ba_{t}$ and apply it to the environment.
}
\label{fig:rnn}
\end{figure}

We formulate learning algorithms as an RL problem and make use of
model-free policy gradient methods to optimize a set parameters
associated with the algorithm. In this setting, the goal is to learn a
policy $\modelp$ that given an observed state $\bs_t$ at step $t$,
estimates a distribution over the next action $\ba_t$, denoted
$\modelp(\ba_t \mid \bs_t)$. Actions represent the commands within the
algorithm and states represent the joint state of the algorithm and
the environment. Previous work in this area has focused on augmenting
a neural network with additional structure and increased
capabilities~\citep{ZarembaS15, Graves16}.  In contrast, we utilize a
simple architecture based on a standard recurrent neural network (RNN)
with LSTM cells~\citep{lstm} as depicted in \figref{fig:rnn}.  At each
episode, the environment is initialized with a latent state $\bphi$,
unknown to the agent, which determines $\bs_1$ and the subsequent
state transition and reward functions.  Once the agent observes
$\bs_1$ as the input to the RNN, the network outputs a distribution
$\modelp(\ba_1\mid\bs_1)$, from which an action $\ba_1$ is sampled.
This action is applied to the environment, and the agent receives a
new state observation $\bs_2$.
The state $\bs_2$ and the previous action $\ba_1$ are then fed into the
RNN and the process repeats until the end of the episode.
%
Upon termination, a reward signal is received.

\section{Learning a Policy by Maximizing Expected Reward}
\label{background}

We start by discussing the most common form of policy gradient,
REINFORCE~\citep{Williams92}, and its entropy regularized
variant~\citep{williams1991function}. REINFORCE has been applied
to model-free policy-based learning with neural networks and
algorithmic domains~\citep{ZarembaS15, Graves16}.



The goal is
to learn a policy $\modelp$ that,
given an
observed state $\bs_t$ at step $t$, estimates a distribution over the
next action $\ba_t$, denoted $\modelp(\ba_t\!\mid\!\bs_t)$.  The
environment is initialized with a latent vector, $\bphi$, which
determines the initial observed state $\bs_1\!=\!\init(\bphi)$, and
the transition function $\bs_{t+1}\!=\!\transition(\bs_t, \ba_t \mid
\bphi)$. 
Note that the use of nondeterministic transitions $\transition$ as
in Markov decision processes (MDP)
may be recovered by assuming that $\bphi$ includes the random seed for 
the any nondeterministic functions.  
Given a latent state $\bphi$, and
$\bs_{1:T}\!\equiv\!(\bs_1,\ldots,\bs_T)$, the model probability of an
action sequence $\ba_{1:T}\!\equiv\!(\ba_1,\ldots,\ba_T)$ is expressed
as,
\begin{equation*}
\modelp(\ba_{1:T} \mid \bphi) = \prod_{t=1}^T \modelp(\ba_t \mid
\bs_t)~,~~~~\text{where}~~~~~\bs_1 = \init(\bphi),~~\bs_{t+1} =
\transition(\bs_t, \ba_t \mid {\bphi})~~\text{for}~~1 \leq t < T~.
\end{equation*}
The environment provides a reward at the end of the episode, denoted
$\reward{\ba_{1:T}}{\bphi}$.  For ease of readability we drop the
subscript from $\ba_{1:T}$ and simply write $\modelp(\bat \mid
\bphi)$ and $\reward{\ba}{\bphi}$.

The objective used to optimize the policy parameters, 
$\btheta$, consists of maximizing expected reward under actions drawn
from the policy, plus an optional maximum entropy regularizer. Given a
distribution over initial latent environment states $p(\bphi)$, we
express the regularized expected reward as,
\begin{equation}
\objrl(\btheta; \temp) = \expected_{\bphi \sim p(\bphi)} \bigg\{
\sum_{\bat \in \A} \modelp(\bat \mid \bphi) \, \Big[ \reward{\bat}{\bphi} - \temp \log \modelp(\bat \mid \bphi) \Big] \bigg\}~.
\label{eq:objrl}
\end{equation}

When $\modelp$ is a non-linear function defined by a neural network,
finding the global optimum of $\btheta$ is challenging, and one often
resorts to gradient-based methods to find a local optimum of
$\objrl(\btheta; \temp)$. Given that
$\frac{\deriv}{\deriv \btheta}\modelp(\ba)
= \modelp(\ba)\frac{\deriv}{\deriv \btheta}\log\modelp(\ba)$ for any
$\ba$ such that $\modelp(\ba) > 0$, one can verify that,
\begin{equation}
\frac{\deriv}{\deriv \btheta}\objrl(\btheta; \temp  \mid \bphi) = 
\sum_{\bat \in \A} \modelp(\bat \mid \bphi) \frac{\deriv}{\deriv \btheta}\log\modelp(\bat \mid \bphi) \, \Big[ \reward{\bat}{\bphi} - \temp \log \modelp(\bat \mid \bphi) -\temp \Big]~.
\label{eq:grad1rl}
\end{equation}
Because the space of possible actions $\A$ is large, enumerating
over all of the actions to compute this gradient is
infeasible. Williams~\yrcite{Williams92} proposed to compute the
stochastic gradient of the expected reward by using Monte Carlo
samples. Using Monte Carlo samples, one first draws $N$ \iid samples
from the latent environment states $\{\bphii\}_{n=1}^N$, and then
draws $K$ \iid samples $\{\bast\}_{k=1}^K$ from
$\modelp(\bat \mid \bphii)$ to approximate the gradient
of \eqref{eq:objrl} by using \eqref{eq:grad1rl} as,
\begin{equation}
\frac{\deriv}{\deriv \btheta}\objrl(\btheta; \temp) \approx \frac{1}{NK}\sum_{n=1}^N
\sum_{k=1}^K \frac{\deriv}{\deriv \btheta}\log \modelp(\bast \mid \bphii) \, \Big[ \reward{\bast}{\bphii} - \temp \log \modelp(\bast \mid \bphii) -\temp \Big].
\label{eq:grad2rl}
\end{equation}
This reparametrization of the gradients is the key to the REINFORCE
algorithm. To reduce the variance of \eqref{eq:grad2rl}, one uses
rewards $\hatrewardd$ that are shifted by some offset values,
\begin{equation}
\hatreward{\bast}{\bphi} = \reward{\bast}{\bphi} - \baseline(\bphi)~,
\end{equation}
where $\baseline$ is known as a {\em baseline} or sometimes called a
{\em critic}. Note that subtracting any offset from the rewards in
\eqref{eq:objrl} simply results in shifting the objective $\objrl$ by
a constant.
\comment{\text{\mohammad{I can't make $\baseline$ function of
$\bs_{1:T}$}}}

Unfortunately, directly maximizing expected reward (\ie~when $\temp =
0$) is prone to getting trapped in a local optimum.  To combat this
tendency, Williams \& Peng~\yrcite{williams1991function} augmented the
expected reward objective by including a maximum entropy regularizer
($\temp > 0$) to promote greater exploration.  We will refer to this
variant of REINFORCE as \pg (maximum entropy exploration).

\section{Under-appreciated Reward Exploration (\rml)}
\label{methodology}

To explain our novel form of policy gradient, we first note that the
optimal policy $\optp$, which globally maximizes $\objrl(\btheta;
\temp \mid \bphi)$ in~\eqref{eq:objrl} for any $\temp > 0$, can be expressed as,
\begin{equation}
\optp(\bat \mid \bphi) = \frac{1}{Z(\bphi)} \exp \Big\{\frac{1}{\temp} \reward{\bat}{\bphi}\Big\}~,
\label{eq:optp}
\end{equation}
where $Z(\bphi)$ is a normalization constant making $\optp$ a
distribution over the space of action sequences $\A$. One can verify
this by first acknowledging that,
\begin{equation}
\objrl(\btheta; \temp \mid \bphi) = -\temp\,\kl{\modelp(\cdot \mid \bphi)}{\optp(\cdot \mid \bphi)}~.
\end{equation}
Since $\kl{p}{q}$ is non-negative and zero iff $p = q$, then $\optp$
defined in \eqref{eq:optp} maximizes $\objrl$. That said, given a
particular form of $\modelp$, finding $\theta$ that exactly characterizes $\optp$ may not be
feasible.

The KL divergence $\kl{\modelp}{\optp}$ is known to be mode seeking
\cite[Section 21.2.2]{murphybook} even with entropy regularization
($\temp > 0$). Learning a policy by optimizing this direction of the
KL is prone to falling into a local optimum resulting in a sub-optimal
policy that omits some of the modes of $\optp$. Although entropy
regularization helps mitigate the issues as confirmed in our
experiments, it is not an effective exploration strategy as it is
undirected and requires a small regularization coefficient $\temp$ to
avoid too much random exploration.  Instead, we propose a directed
exploration strategy that improves the mean seeking behavior of policy
gradient in a principled way.


We start by considering the alternate mean seeking direction of the KL
divergence, $\kl{\optp}{\modelp}$. \cite{rml2016} considered this
direction of the KL to directly learn a policy by optimizing
\begin{equation}
\objrml(\btheta; \temp) = 
\expected_{\bphi \sim p(\bphi)} \bigg\{ {\temp} \sum_{\bat \in \A}
\,\optp(\bat \mid \bphi)
\log \modelp(\bat \mid \bphi) \bigg\}~,
\label{eq:objrml}
\end{equation}
for structured prediction.  This objective has the same optimal
solution $\optp$ as $\objrl$ since,
\begin{equation}
\objrml(\btheta; \temp \mid \bphi) 
= 
-\temp\,\kl{\optp(\cdot \mid \bphi)}{\modelp(\cdot \mid \bphi)}
+\text{const}
~.
\end{equation}
\citet{rml2016} argue that in some structured prediction problems when
one can draw samples from $\optp$, optimizing \eqref{eq:objrml} is
more effective than \eqref{eq:objrl}, since no sampling from a
non-stationary policy $\modelp$ is required. If $\modelp$ is a
log-linear model of a set of features, $\objrml$ is convex in
$\btheta$ whereas $\objrl$ is not, even in the log-linear case.
Unfortunately, in scenarios that the reward landscape is unknown or
computing the normalization constant $Z(\bphi)$ is intractable,
sampling from $\optp$ is not straightforward.

In RL problems, the reward landscape is completely unknown, hence
sampling from $\optp$ is intractable. 
\comment{
One remedy to this is to alter the objective to make it more 
amenable to sampling using $\modelp$.  This is evident
in the objective given by Reward-Weighted Regression (RWR)~\citep{rwr},
which has been used to tackle RL problems in the past:
\begin{equation}
\objrwr(\btheta; \temp \mid \bphi) = \expected_{\bphi \sim p(\bphi)} 
\bigg\{ \sum_{\bat \in \A} \optp(\bat \mid \bphi) \modelp(\bat\mid\bphi)
\bigg\}~.
\end{equation}

In contrast, this paper leaves the form of \eqref{eq:objrml} intact
and instead proposes to
}
This paper proposes to
approximate the expectation with respect to $\optp$
by using {\em self-normalized importance
  sampling}~\citep{mcbook}, where the proposal distribution is
$\modelp$ and the reference distribution is $\optp$. For importance
sampling, one draws $K$ \iid samples $\{\bast\}_{k=1}^K$ from
$\modelp(\bat \mid \bphi)$ and computes a set of normalized importance
weights to approximate $\objrml(\btheta; \temp \mid \bphi)$ as,
\begin{equation}
\objrml(\btheta; \temp \mid \bphi) \approx \temp
\sum_{k=1}^K \frac{\imw(\bast \mid \bphi)}{\sum_{m=1}^K\imw(\bat^{(m)} \mid \bphi)} \, \log\modelp(\bast \mid \bphi)~,
\label{eq:is-rml}
\end{equation}
where $\imw(\bast \mid \bphi) \propto \optp / \modelp$ denotes an
importance weight defined by,
\begin{equation}
\imw(\bast \mid \bphi) = \exp \Big\{\frac{1}{\temp} \reward{\bast}{\bphi}
- \log\modelp(\bast \mid \bphi)\Big\}~.
\end{equation}
One can view these importance weights as evaluating the discrepancy between scaled
rewards $r / \temp$ and the policy's log-probabilities $\log
\modelp$. Among the $K$ samples, a sample 
that is least appreciated by the model, \ie~has the largest $r / \temp
- \log \modelp$, receives the largest positive feedback in
\eqref{eq:is-rml}.

In practice, we have found that just using the importance sampling RAML
objective in~\eqref{eq:is-rml} does not always yield promising
solutions.  Particularly, at the beginning of training, when $\modelp$
is still far away from $\optp$, the variance of importance weights is
too large, and the self-normalized importance sampling procedure
results in poor approximations.
To stabilize early phases of training and ensure that the model
distribution $\modelp$ achieves large expected reward scores, we
combine the expected reward and RAML objectives to benefit from the
best of their mode and mean seeking behaviors. Accordingly, we propose
the following objective that we call {\em under-appreciated reward
  exploration~(\rlrml)},
\begin{equation}
\objrlrml(\btheta; \temp) = \expected_{\bphi \sim p(\bphi)} \bigg\{
\sum_{\bat \in \A} \Big[\modelp(\bat \mid \bphi) \, \reward{\bat}{\bphi} + 
\temp\,\optp(\bat \mid \bphi)
\log \modelp(\bat \mid \bphi) \Big] \bigg\}~,
\label{eq:objrml2}
\end{equation}
which is the sum of the expected reward and RAML objectives. In our
preliminary experiments, we considered a composite objective of
$\objrl + \objrml$, but we found that removing the entropy term is
beneficial. Hence, the $\objrlrml$ objective does not include entropy
regularization. Accordingly, the optimum policy for $\objrlrml$ is no
longer $\optp$, as it was for $\objrl$ and
$\objrml$. Appendix~\ref{sec:appxB} derives the optimal policy for
$\objrlrml$ as a function of the optimal policy for $\objrl$. We find
that the optimal policy of \rml is more sharply concentrated on the
high reward regions of the action space, which may be an advantage for
\rml, but we leave more analysis of this behavior to future work.

\comment{
as ${\optp(\bat\mid\bphi)} / {(\alpha(\bphi) -
  \reward{\bat}{\bphi} / \temp)}$, where $\alpha(\bphi) >
\max (\reward{\bat}{\bphi} / \temp)$ normalizes the distribution for
each $\bphi$.
}

To compute the gradient of $\objrlrml(\btheta; \temp)$, we use the
self-normalized importance sampling estimate outlined in
\eqref{eq:is-rml}.  We assume that the importance weights are constant
and contribute no gradient to $\frac{\deriv}{\deriv
  \btheta}\objrlrml(\btheta; \temp)$. To approximate the gradient, one
draws $N$ \iid samples from the latent environment states
$\{\bphii\}_{n=1}^N$, and then draws $K$ \iid samples
$\{\bast\}_{k=1}^K$ from $\modelp(\bat\!\mid\!\bphii)$ to obtain
\begin{equation}
\frac{\deriv}{\deriv \btheta}\objrlrml(\btheta; \temp) \approx
\!\frac{1}{N}\!\sum_{n=1}^N
\sum_{k=1}^K \frac{\deriv}{\deriv \btheta}\log \modelp(\bast\!\mid\!\bphii)\!\bigg[\frac{1}{K}\hatreward{\bast}{\bphii} + \temp\frac{\imw(\bast\!\mid\!\bphii)}{\sum_{m=1}^K\!\!\imw(\bat^{(m)}\!\mid\!\bphii)} \!\bigg].
\label{eq:gradrlrml}
\end{equation}
As with REINFORCE, the rewards are shifted by an offset $b(\bphi)$.
In this gradient, the model log-probability of a sample action
sequence $\bast$ is reinforced if the corresponding reward is large,
or the corresponding importance weights are large, meaning that the
action sequence is under-appreciated. The normalized importance
weights are computed using a softmax operator $\softmax(r / \temp -
\log \modelp)$.

\comment{
There are three ways to explain why $\modelp$ may not converge to
$\optp$ after training using PG:
\begin{enumerate}
\item{\bf Model capacity} -- The form of the model may not be expressive enough to represent
$\optp$.
\item{\bf Variance} -- The space of possible actions $\A$ is very
large and the reward is very sparse causing the stochastic gradients
defined in \eqref{eq:grad2rl} to have a large variance.
\item{\bf Optimization} -- The parameters may fall into local
optima of the objective, or the objective may have a very slow
convergence rate.
\end{enumerate}
When using recurrent neural networks for simple tasks, point (1)
should not be an issue. Usually papers blame variance for the failure
cases of PG, as it is hard to reason about optimization landscape
especially in high-dimensional problems. Below we discuss a very
simple one-dimensional problem where local optima seem to be a problem
for the regularized expected reward objective. \mohammad{find a simple
example!}
}

\comment{
\begin{figure}[t]
\begin{center}
\begin{tikzpicture} 
\begin{axis}[
    width=16cm,xlabel={$\theta$},
    ylabel={},axis x line=middle, axis y line=middle, 
    title={}] 
\addplot[red,domain=-6:6, no markers, samples=100]
{1/(1+exp(x)) * ln(1/(1+exp(x))) + exp(x)/(1+exp(x)) * ln(exp(x)/(1+exp(x)))
-(1/(1+exp(x))) * ln(.05) + -(exp(x)/(1+exp(x))) * ln(.95)}; 
\addplot[blue,domain=-6:6, no markers, samples=100]
{.05 * ln(.05) + .95 * ln(.95) - .05 * ln(1/(1+exp(x))) - .95 * ln(exp(x)/(1+exp(x))) }; 
\addplot[green,domain=-6:6, no markers, samples=100]
{.05 * ln(.05) + .95 * ln(.95)
- .05 * ln(1/(1+exp(x))) - .95 * ln(exp(x)/(1+exp(x)))
+1/(1+exp(x)) * ln(1/(1+exp(x))) + exp(x)/(1+exp(x)) * ln(exp(x)/(1+exp(x)))
-(1/(1+exp(x))) * ln(.05) + -(exp(x)/(1+exp(x))) * ln(.95)}; 
\addplot[orange,domain=-6:6, no markers, samples=100]
{.05 * ln(.05) + .95 * ln(.95)
- .05 * ln(1/(1+exp(x))) - .95 * ln(exp(x)/(1+exp(x)))
-(1/(1+exp(x))) * ln(.05) + -(exp(x)/(1+exp(x))) * ln(.95)}; 
\legend{$\kl{\modelp}{\optp}$,
$\kl{\optp}{\modelp}$,
$\kl{\optp}{\modelp} + \kl{\modelp}{\optp}$,
$\kl{\modelp}{\optp} + \kl{\modelp}{\optp} - \ent{\modelp}$}
\end{axis} 
\end{tikzpicture}
\end{center}
\end{figure}
}

\comment{
\begin{figure}[t]
\begin{center}
\begin{tikzpicture} 
\begin{axis}[
    width=16cm,xlabel={$\theta$},
    ylabel={},axis x line=middle, axis y line=middle, 
    title={}] 
\addplot[red,domain=-15:15, no markers, samples=100]
{1/(1+exp(x)) * ln(1/(1+exp(x))) + exp(x)/(1+exp(x)) * ln(exp(x)/(1+exp(x)))
-(1/(1+exp(x))) * ln(.2) + -(exp(x)/(1+exp(x))) * ln(.8)
+exp(-x)/(exp(-x)+exp(-2*x))   * ln(exp(-x)/(exp(-x)+exp(-2*x)))
+exp(-2*x)/(exp(-x)+exp(-2*x)) * ln(exp(-2*x)/(exp(-x)+exp(-2*x)))
-exp(-x)/(exp(-x)+exp(-2*x))   * ln(.5) + -exp(-2*x)/(exp(-x)+exp(-2*x)) * ln(.5)}; 
\legend{$\kl{\modelp}{\optp}$}
\end{axis} 
\end{tikzpicture}
\end{center}
\end{figure}
}

\comment{
We use recurrent neural networks (RNNs) to model $\btheta$.
~\cite{williams1991function\extracite{,mnih2016asynchronous}}), which
is formulated as minimization of the following objective,

where $\reward{\byp}{\bys}$ denotes the reward function, \eg~negative
edit distance or BLEU score, $\temp$ controls the degree of
regularization, and $\ent{p}$ is the entropy of a distribution $p$,
\ie~$\ent{p(\byp)} = -\sum_{\byp \in \calY}p(\byp) \log p(\byp)$. It
is well-known that optimizing $\objrl(\btheta; \temp)$ using SGD is
challenging because of the large variance of the gradients. Below we
describe how ML and RL objectives are related, and propose a hybrid
between the two that combines their benefits for supervised learning.
}

\vspace{-.1cm}
\section{Related Work}
\vspace{-.1cm}
\label{related_work}



Before presenting the experimental results, we briefly review some
pieces of previous work that closely relate to the \rml approach.

\textbf{Reward-Weighted Regression.\,}
Both RAML and \rml objectives bear some similarity to a method in
continuous control known as Reward-Weighted Regression
(RWR)~\citep{rwr, rwr2}. Using our notation, the RWR objective is
expressed as,
\begin{eqnarray}
\label{eq:rwr}
\objrwr(\btheta; \temp \mid \bphi) &=& 
\log\sum_{\bat \in \A} \optp(\bat \mid \bphi) \modelp(\bat\mid\bphi)\\
\label{eq:vl-rwr}
&\geq& 
\sum_{\bat \in \A} q(\bat \mid \bphi) \log\frac{\optp(\bat\mid\bphi)\modelp(\bat\mid\bphi)}{q(\bat \mid \bphi)}~.
\end{eqnarray}
To optimize $\objrwr$, \citet{rwr} propose a technique inspired by the
EM algorithm to maximize a variational lower bound
in \eqref{eq:vl-rwr} based on a variational distribution
$q(\bat \mid \bphi)$. The RWR objective can be interpreted as a log of
the correlation between $\optp$ and $\modelp$. By contrast, the RAML
and \rml objectives are both based on a KL divergence between $\optp$
and $\modelp$.

To optimize the RWR objective, one formulates the gradient as,
\begin{equation}
\label{eq:drwr1}
\frac{\deriv}{\deriv \btheta} \objrwr(\btheta; \temp \mid \bphi) =
\sum_{\bat \in \A} \frac{\optp(\bat \mid \bphi) \modelp(\bat \mid \bphi)}{C} \frac{\deriv}{\deriv \btheta} \log \modelp(\bat \mid \bphi),
\end{equation}
where $C$ denotes the normalization factor, \ie~$C =
{\sum_{\bat \in \A} \optp(\bat \mid \bphi) \modelp(\bat \mid \bphi)}$.
The expectation with respect to
$\optp(\bat \mid \bphi)\modelp(\bat \mid \bphi) / C$ on the RHS can be
approximated by self-normalized importance
sampling,\footnote{\citet{bornschein2014reweighted} apply the same
trick to optimize the log-likelihood of latent variable models.}
where
the proposal distribution is $\modelp$. Accordingly, one draws $K$
Monte Carlo samples $\{\bast\}_{k=1}^K$ \iid from
$\modelp(\bat\!\mid\!\bphi)$ and formulates the gradient as,
\begin{equation}
\label{eq:drwr2}
\frac{\deriv}{\deriv \btheta} \objrwr(\btheta; \temp \mid \bphi)
\approx \frac{1}{K}\sum_{k=1}^K 
\frac{u(\bast \mid \bphi)}{\sum_{m=1}^K u(\bat^{(m)} \mid \bphi)}
\frac{\deriv}{\deriv \btheta}\log \modelp(\bast\mid\bphi),
\end{equation}
where $u(\bast \mid \bphi)
= \exp\{\frac{1}{\temp} \reward{\bast}{\bphi} \}$. There is some
similarity between~\eqref{eq:drwr2} and~\eqref{eq:is-rml} in that they
both use self-normalized importance sampling, but note the critical
difference that \eqref{eq:drwr2} and~\eqref{eq:is-rml} estimate the
gradients of two different objectives, and hence the importance
weights in~\eqref{eq:drwr2} do not correct for the sampling
distribution $\modelp(\bat\!\mid\!\bphi)$ as opposed
to~\eqref{eq:is-rml}.

Beyond important technical differences, the optimal policy of
$\objrwr$ is a one hot distribution with all probability mass concentrated
on an action sequence with maximal reward, whereas the optimal policies
for RAML and UREX are everywhere nonzero,
with the probability of different action sequences being assigned proportionally
to their exponentiated reward (with UREX introducing an additional re-scaling;
see Appendix~\ref{sec:appxB}).  Further, the notion of under-appreciated
reward exploration evident in $\objrlrml$, which is key to \rml's
performance, is missing in the RWR formulation.

\textbf{Exploration.\,}
The RL literature contains many different attempts at incorporating
exploration that may be compared with our method.
The most common exploration strategy considered in value-based RL is
$\epsilon$-greedy Q-learning, where at each step the agent either
takes the best action according to its current value approximation or
with probability $\epsilon$ takes an action sampled uniformly at
random.  Like entropy regularization, such an approach applies
undirected exploration, but it has achieved recent success in game
playing environments~\citep{atari, dqn, mnih2016asynchronous}.

Prominent approaches to improving exploration beyond $\epsilon$-greedy
in value-based or model-based RL have focused on reducing uncertainty
by prioritizing exploration toward states and actions where the agent
knows the least.  This basic intuition underlies work on counter and
recency methods~\citep{thrun92}, exploration methods based on
uncertainty estimates of values~\citep{interval_estimation,
tokic2010adaptive}, methods that prioritize learning environment
dynamics \citep{e3, stadie_expl}, and methods that provide an
intrinsic motivation or curiosity bonus for exploring unknown
states~\citep{curiosity, intrinsic}. 

In contrast to value-based methods, exploration for policy-based RL
methods is often a by-product of the optimization algorithm itself.
Since algorithms like REINFORCE and Thompson sampling choose actions
according to a stochastic policy, sub-optimal actions are chosen with
some non-zero probability.  The Q-learning algorithm may also be
modified to sample an action from the softmax of the Q values rather
than the argmax~\citep{suttonbook}.

Asynchronous training has also been reported to have an exploration
effect on both value- and policy-based methods.
\citet{mnih2016asynchronous} report that asynchronous training 
can stabilize training by reducing the bias experienced by a single trainer.
By using multiple separate trainers, an agent is less likely 
to become trapped at a policy found to be locally optimal only
due to local conditions.  In the same spirit, \citet{bootstrap_dqn}
use multiple Q value approximators and sample only one to act for each
episode as a way to implicitly incorporate exploration.

By relating the concepts of value and policy in RL,
the exploration strategy we propose tries to bridge the 
discrepancy between the two.
In particular, UREX can be viewed as a hybrid combination of 
value-based and policy-based exploration strategies
that attempts to capture the benefits of each.

\textbf{Per-step Reward.\,}
Finally, while we restrict ourselves to episodic settings where a
reward is associated with an entire episode of states and actions,
much work has been done to take advantage of environments that provide
per-step rewards.  These include policy-based methods such as
actor-critic~\citep{mnih2016asynchronous, schulmaniclr2016} and
value-based approaches based on Q-learning~\citep{dqn, pdqn}.  Some of
these value-based methods have proposed a softening of Q-values which
can be interpreted as adding a form of maximum-entropy
regularizer~\citep{mellowmax, azar, fox, ziebart2010modeling}.  The
episodic total-reward setting that we consider is naturally harder
since the credit assignment to individual actions within an episode is
unclear.

\comment{
Although the problem of program induction from examples has been studied in many fields,
including program induction, inductive logic programming \citep{ilpbook},
relational learning \citep{kempetal07} and
regular language learning
\citep{angulin87};
we do not attempt 
a comprehensive survey here.
Instead, we focus on related neural network approaches 
for solving algorithmic tasks
and relevant work that considers exploration in reinforcement learning.


\paragraph{Learning Algorithms.}
Previous work on neural network learning algorithms 
for algorithmic problems
falls into two general 
categories: strongly supervised maximum likelihood training that 
\emph{learns from examples} using completely known input-output pairs; 
and self-driven training via reinforcement learning that 
\emph{learns from acting}, 
where the correct output is not known
but a reward signal is available to indicate which outputs are preferable
to others.  

The former approach imposes more constraints than the latter and therefore has had more success.
While some supervised learning approaches use a simple sequence-to-sequence 
recurrent neural network to map inputs to outputs \citep{ZarembaS14}, 
more successful approaches have been based on augmenting the model with 
additional computational elements that have been modified 
to allow for differentiation via backpropagation.  
For example, \cite{GravesWD14,Graves16} have proposed the Neural Turing Machine 
and the Differentiable Neural Computer (DNC), 
both of which provide an auxiliary memory with a trainable 
interface for reading and writing.  
Beyond basic sequence manipulations, these models have also 
successfully learned tasks such as finding the shortest path between 
specified points.  
Others have investigated stack- or queue-based augmentations
\citep{Grefenstette15, JoulinM15},
which have been successfully applied to simple sequence manipulation tasks, 
and simple arithmetic problems, such as integer addition.  

Other supervised neural network approaches have not augmented the network 
but rather altered how it manipulates or interacts with the input.  
Two prominent examples are the Grid LSTM \citep{KalchbrennerDG15} 
and the Neural GPU \citep{KaiserS15}, which have succeeded in solving 
addition and multiplication problems given variable-length inputs.  
A recent supervised approach that does not learn from input-output pairs 
but rather from full execution traces is given by \cite{ReedF15}.  
By providing much richer training data, the network is able to learn more 
complex algorithms like sorting.

All of these approaches require strong supervision that provides, at a minimum,
the correct output for any given input.
There are many cases where this is undesirable.  
For example, in an NP-complete problem such as the Traveling Salesman Problem,
the correct output is infeasible to compute for large inputs,
while a proposed output is easy to check.
Similarly, in tasks where one cares about the efficiency of the algorithm 
produced in addition to correctness
(for example in binary search we prefer a logarithmic-time algorithm 
over a linear-time algorithm),
simply providing 
the correct output for a given input does not convey sufficient information.
Yet, providing the model with a reward that indicates
the cost of its execution is straightforward.

Therefore, a more general and potentially more desirable approach
is to deploy self-driven learning that only receives reinforcement
based on its outputs.
Unfortunately, such an approach has experienced less success in practice,
since the model receives less information during training,
which has also resulted in far less research activity to date.
However, some recent work has considered a reinforcement based approach
to algorithmic tasks,
including \citep{ZarembaS15}, where a Neural Turing Machine 
is trained via reinforcement learning.  
The resulting model can solve simple sequence manipulation tasks such as 
deduplicating or reversing a sequence.  
Also, previous work on the DNC \citep{Graves16}
has also briefly explored training via reinforcement learning, 
presenting the ability of the DNC to solve a block manipulation game 
given only reward feedback.


\paragraph{Exploration in Reinforcement Learning.}
}

\section{Six Algorithmic Tasks}

We assess the effectiveness of the proposed approach on five algorithmic tasks
from the OpenAI Gym \citep{Brockman},
as well as a new {\em binary search} problem.
Each task is summarized below, with further details available on
the Gym website\footnote{\url{gym.openai.com}} or in the
corresponding
open-source code.\footnote{\url{github.com/openai/gym}}
In each case, the environment has a hidden tape and a hidden sequence. 
The agent observes the sequence via a pointer to a single character, which
can be moved by a set of {\em pointer control actions}. 
Thus an action  $\ba_t$ is represented as a tuple $(m, w, o)$ where $m$
denotes how to move, $w$ is a boolean denoting whether to write, and
$o$ is the output symbol to write.
\begin{enumerate}[topsep=0em,itemsep=.5em,leftmargin=1.3em,parsep=0em]
\item \textbf{Copy}: The agent should emit a copy of the
  sequence. The pointer actions are move left and right.
\item \textbf{DuplicatedInput}: In the hidden tape, each character is
  repeated twice. The agent must deduplicate the sequence and emit
  every other character. The pointer actions are move left and right.
\item \textbf{RepeatCopy}: The agent should emit the hidden sequence
  once, then emit the sequence in the reverse order, then emit the
  original sequence again. The pointer actions are move left and
  right.
\item \textbf{Reverse}: The agent should emit the hidden sequence in
  the reverse order. As before, the pointer actions are move left and
  right.
\item \textbf{ReversedAddition}: The hidden tape is a $2\!\times\!n$
  grid of digits representing two numbers in base $3$ in little-endian
  order. The agent must emit the sum of the two numbers, in
  little-endian order.  The allowed pointer actions are move left,
  right, up, or down.
\end{enumerate}
The OpenAI Gym provides an additional harder task called
ReversedAddition3, which involves adding three numbers. We omit this
task, since none of the methods make much progress on it.

For these tasks, the input sequences encountered during training
range from a length of $2$ to $33$ characters. 
A reward of $1$ is given for each correct
emission.  On an incorrect emission, a small penalty of $-0.5$ is
incurred and the episode is terminated. The agent is also terminated
and penalized with a reward of $-1$ if the episode exceeds a certain
number of steps. 
For the experiments using \rml and \pg, we associate an episodic sequence of actions
with the \emph{total} reward, defined as the sum of the per-step rewards.  
The experiments using Q-learning, on the other hand, used the per-step rewards.
Each of the Gym tasks has a \emph{success threshold},
which determines the required average reward over $100$ episodes for
the agent to be considered successful.

We also conduct experiments on an additional algorithmic task
described below:
\begin{enumerate}[topsep=-.2em,itemsep=0em,leftmargin=1.3em,parsep=-.2em]
  \setcounter{enumi}{5}
\item \textbf{BinarySearch}: Given an integer $n$, the environment has
  a hidden array of $n$ distinct numbers stored in ascending
  order. The environment also has a query number $x$ unknown to the
  agent that is contained somewhere in the array. The goal of the
  agent is to find the query number in the array in a small number
  of actions. The environment has three integer registers initialized
  at $(n, 0, 0)$.  At each step, the agent can interact with the
  environment via the four following actions:\vspace*{.3em}
  \begin{itemize}[topsep=0em,itemsep=.2em,leftmargin=1.2em,parsep=0em]
    \setcounter{enumi}{5}
  \item $\mathrm{INC}(i)$: increment the value of the register $i$ for $i \in \{1,\, 2,\, 3\}$.
  \item $\mathrm{DIV}(i)$: divide the value of the register $i$ by 2 for $i \in \{1,\, 2,\, 3\}$.
  \item $\mathrm{AVG}(i)$: replace the value of the register $i$ with
    the average of the two other registers.
  \item $\mathrm{CMP}(i)$: compare the value of the register $i$ with
    $x$ and receive a signal indicating which value is greater. The
    agent succeeds when it calls $\mathrm{CMP}$ on an array cell
    holding the value $x$.
  \end{itemize}\vspace*{.5em}
  The agent is terminated when the number of steps exceeds a maximum
  threshold of $\bsmax$ steps and recieves a reward of $0$. If the
  agent finds $x$ at step $t$, it recieves a reward of $10(1 - {t} /
  {(\bsmax)})$.
\end{enumerate}
We set the maximum number of steps to $\bsmax$ to allow the agent to
perform a full linear search. A policy performing full linear search
achieves an average reward of $5$, because $x$ is chosen uniformly at
random from the elements of the array.  A policy employing binary
search can find the number $x$ in at most $2\log_2 n + 1$ steps. If
$n$ is selected uniformly at random from the range $32\le n\le 512$,
binary search yields an optimal average reward above $9.55$.  We set
the \emph{success threshold} for this task to an average reward of
$9$.

%

\vspace*{-.1cm}
\section{Experiments}
\vspace*{-.1cm}
\label{experimental_setup}

We compare our policy gradient method using under-appreciated reward
exploration (\rml) against two main RL baselines: (1) REINFORCE with
entropy regularization termed \pg \citep{williams1991function},
where the value of $\temp$ determines the degree of
regularization. When $\temp=0$, standard REINFORCE is obtained. (2)
one-step double Q-learning based on bootstrapping one step future
rewards.

\vspace*{-.1cm}
\subsection{Robustness to hyper-parameters}
\vspace*{-.1cm}

Hyper-parameter tuning is often tedious for RL algorithms. We found
that the proposed \rml method significantly improves robustness
to changes in hyper-parameters when compared to \pg.
For our
experiments, we perform a careful grid search over a set of
hyper-parameters for both \pg and \rml. For any hyper-parameter
setting, we run the \pg and \rml methods $5$ times with different random
restarts. We explore the following main hyper-parameters:
\begin{itemize}[topsep=0em,itemsep=.2em,leftmargin=2em,parsep=0em]
\item The {\em learning rate} denoted $\eta$ chosen from a set of $3$
  possible values $\eta\in\{0.1, 0.01, 0.001\}$.
\item The maximum L2 norm of the gradients, beyond which the
  gradients are {\em clipped}. This parameter, denoted $c$, matters
  for training RNNs. The value of $c$ is selected from $c\in\{1,
  10, 40, 100\}$.
\item The temperature parameter $\tau$ that controls the degree of
  exploration for both \pg and \rml. For \pg, we use $\tau\in\{0,
  0.005, 0.01, 0.1\}$.  For \rml, we only consider $\tau=0.1$, which
  consistently performs well across the tasks.
\end{itemize}
In all of the experiments, both \pg and \rml are treated exactly the
same. In fact, the change of implementation is just a few lines of
code. Given a value of $\tau$, for each task, we run $60$ training
jobs comprising $3$ learning rates, $4$ clipping values, and $5$
random restarts. We run each algorithm for a maximum number of steps
determined based on the difficulty of the task. The training jobs for
Copy, DuplicatedInput, RepeatCopy, Reverse, ReversedAddition, and
BinarySearch are run for $2K$, $500$, $50K$, $5K$, $50K$, and $2K$
stochastic gradient steps, respectively. We find that running a
trainer job longer does not result in a better performance. Our policy
network comprises a single LSTM layer with $128$ nodes. We use the
Adam optimizer~\citep{adam} for the experiments.

\begin{table}[t]
\caption{Each cell shows the percentage of $60$ trials with
different hyper-parameters ($\eta$, $c$) and random restarts that
successfully solve an algorithmic task. \rml is more robust to
hyper-parameter changes than \pg. We evaluate \pg with a few
temperatures and \rml with $\temp\!=\!0.1$.\vspace*{.1cm}}
\label{gym-results}
\begin{center}
\begin{tabular}{c|c|c|c|c|c|}
\cline{2-6}
 & \multicolumn{4}{c|}{REINFORCE~/~\pg} & \multicolumn{1}{c|}{\rml} \\ \cline{2-6}
 & \multicolumn{1}{c|}{$\tau=0.0$} & \multicolumn{1}{c|}{$\tau=0.005$} & \multicolumn{1}{c|}{$\tau=0.01$} & 
   \multicolumn{1}{c|}{$\tau=0.1$} & \multicolumn{1}{c|}{$\tau=0.1$} \\ \hline
\multicolumn{1}{|l|}{Copy} 		&  85.0 & 88.3 & {\bf 90.0} & 3.3 & 75.0 \\ \hline
\multicolumn{1}{|l|}{DuplicatedInput} 	& 68.3 & 73.3 & 73.3 & 0.0 & {\bf 100.0} \\ \hline
\multicolumn{1}{|l|}{RepeatCopy} 	& 0.0 & 0.0 & 11.6 & 0.0 & {\bf 18.3} \\ \hline
\multicolumn{1}{|l|}{Reverse} 		& 0.0 & 0.0 & 3.3 & 10.0 & {\bf 16.6} \\ \hline
\multicolumn{1}{|l|}{ReversedAddition} 	& 0.0 & 0.0 & 1.6 & 0.0 & {\bf 30.0} \\ \hline
\multicolumn{1}{|l|}{BinarySearch} 	& 0.0 & 0.0 & 1.6 & 0.0 & {\bf 20.0} \\ \hline
\end{tabular}
\end{center}
\vspace*{-.2cm}
\end{table}

\tabref{gym-results} shows the percentage of $60$ trials on
different hyper-parameters ($\eta$, $c$) and random restarts which
successfully solve each of the algorithmic tasks. It is clear that
\rml is more robust than \pg to changes in hyper-parameters,
even though we only report the results of \rml for a single
temperature. See Appendix \ref{appendix} for more detailed tables on
hyper-parameter robustness.

\vspace*{-.1cm}
\subsection{Results}
\vspace*{-.1cm}

\begin{figure}[h]
\begin{center}
\includegraphics[width=1.0\textwidth]{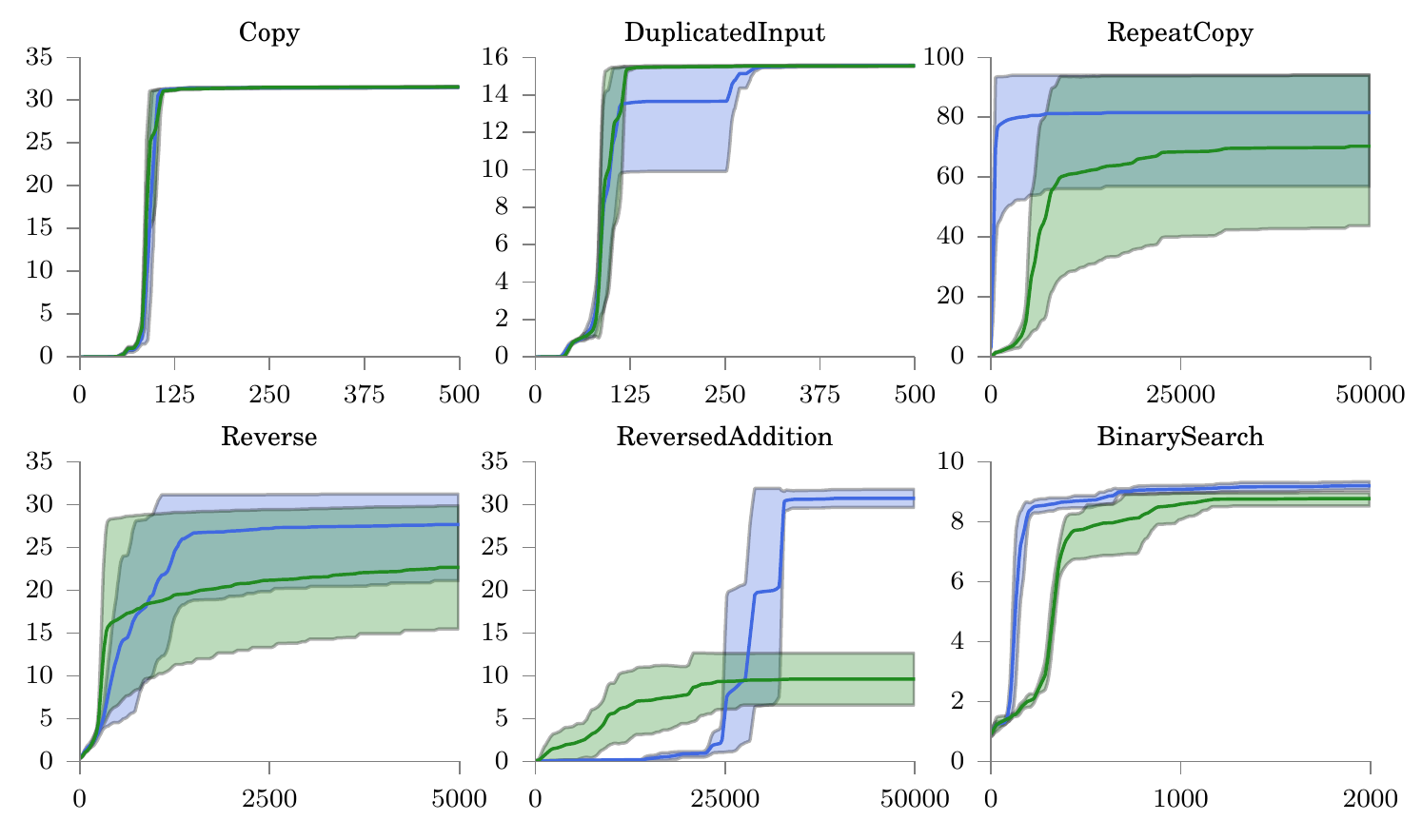}
\end{center}
\caption{Average reward during training for \pg (green) and \rml (blue).  
We find the best hyper-parameters for each method, and run each algorithm
$5$ times with random restarts.  The curves present the average reward as 
well as the single standard deviation region clipped at the min and max.}
\label{fig:rewards}
\end{figure}

\tabref{gym-results2} presents the number of successful attempts (out
of $5$ random restarts) and the expected reward values (averaged over
$5$ trials) for each RL algorithm given the best
hyper-parameters. One-step Q-learning results are also included in the
table. We also present the training curves for \pg and \rml in
\figref{fig:rewards}.
It is clear that \rml outperforms the baselines on these
tasks. On the more difficult tasks, such as Reverse and
ReverseAddition, \rml is able to consistently find an appropriate
algorithm, but \pg and Q-learning fall behind. Importantly, for the
BinarySearch task, which exhibits many local maxima and necessitates
smart exploration,
\rml is the only method that can solve it consistently.  
The Q-learning baseline solves some of the simple tasks, but it makes
little headway on the harder tasks. We believe that entropy
regularization for policy gradient and $\epsilon$-greedy for
Q-learning are relatively weak exploration strategies in long episodic
tasks with delayed rewards. On such tasks, one random exploratory step
in the wrong direction can take the agent off the optimal policy,
hampering its ability to learn.  In contrast, \rml provides a form of
adaptive and smart exploration.  In fact, we observe that the variance
of the importance weights decreases as the agent approaches the
optimal policy, effectively reducing exploration when it is no longer
necessary; see Appendix~\ref{appendix-variance}.

\begin{table}[t]
\caption{Results on several algorithmic tasks comparing Q-learning and
policy gradient based on \pg and \rml. We find the best
hyper-parameters for each method, and run each algorithm $5$ times
with random restarts. Number of successful attempts (out of $5$) that
achieve a reward threshold is reported. Expected reward computed over
the last few iterations of training is also reported.}
\label{gym-results2}
\begin{center}
\begin{tabular}{c|c|c|c||c|c|c|c}
\cline{2-7}
& \multicolumn{3}{c||}{Num. of successful attempts out of $5$} & \multicolumn{3}{c|}{Expected reward}\\
\cline{2-7}
 & \multicolumn{1}{c|}{Q-learning} & \multicolumn{1}{c|}{~~\pg~~} & \multicolumn{1}{c||}{~~\rml~~}
 & \multicolumn{1}{c|}{Q-learning} & \multicolumn{1}{c|}{~~\pg~~} & \multicolumn{1}{c|} {~~\rml~~}\\ \hline
\multicolumn{1}{|l|}{Copy}
                                       & 5    & 5    & 5
                                       & 31.2 & 31.2 & 31.2
                                       \\ \hline
\multicolumn{1}{|l|}{DuplicatedInput}
                                       & 5    & 5    & 5
                                       & 15.4 & 15.4 & 15.4
                                       \\ \hline
\multicolumn{1}{|l|}{RepeatCopy}
                                       & 1    & 3    & 4
                                       & 39.3 & 69.2 & 81.1
                                       \\ \hline
\multicolumn{1}{|l|}{Reverse}
                                       & 0    & 2    & 4
                                       & 4.4  & 21.9 & 27.2
                                       \\ \hline
\multicolumn{1}{|l|}{ReversedAddition}
                                       & 0    & 1    & 5
                                       & 1.1  & 8.7  & 30.2
                                       \\ \hline
\multicolumn{1}{|l|}{BinarySearch}
                                       & 0    & 1    & 4
                                       & 5.2  & 8.6  & 9.1
                                       \\ \hline
\end{tabular}
\end{center}
\end{table}

\subsection{Generalization to longer sequences}
To confirm whether our method is able to find the correct algorithm
for multi-digit addition, we investigate its generalization to longer
input sequences than provided during training. We evaluate the trained
models on inputs up to a length of $2000$ digits, even though training
sequences were at most $33$ characters. For each length, we test the
model on $100$ randomly generated inputs, stopping when the accuracy
falls below $100\%$. Out of the $60$ models trained on addition with
\rml, we find that $5$ models generalize to numbers up to $2000$
digits without any observed mistakes. On the best \rml
hyper-parameters, $2$ out of the $5$ random restarts are able to
generalize successfully. For more detailed results on the
generalization performance on $3$ different tasks including Copy,
DuplicatedInput, and ReversedAddition, see
Appendix~\ref{appendix-generalization}.  During these evaluations, we
take the action with largest probability from $\modelp(\bat\mid\bphi)$
at each time step rather than sampling randomly.

We also looked into the generalization of the models trained on the
BinarySearch task. We found that none of the agents perform proper
binary search. Rather, those that solved the task perform a hybrid of
binary and linear search: first actions follow a binary search
pattern, but then the agent switches to a linear search procedure once
it narrows down the search space; see Appendix~\ref{appendix-traces}
for some execution traces for BinarySearch and
ReversedAddition. Thus, on longer input sequences, the agent's running
time complexity approaches linear rather than logarithmic. We hope
that future work will make more progress on this task. This task is
especially interesting because the reward signal should incorporate
both correctness and efficiency of the algorithm.

\comment{
This task is interesting because there exist many local maxima in the
policy space.  Even between a full linear search and an efficient
binary search there are many possible policies that solve the task.
As we will see later, none of our methods are able to find the
intended efficient binary search algorithm, but rather at best find
one of these policies in between linear and binary search.  We hope
that future work in this area can make more headway on this difficult
task.
}

\subsection{Implementation details}

In all of the experiments, we make use of curriculum learning.  The
environment begins by only providing small inputs and moves on to
longer sequences once the agent achieves close to maximal reward over
a number of steps. For policy gradient methods including \pg
and \rml, we only provide the agent with a reward at the end of the
episode, and there is no notion of intermediate reward. For the
value-based baseline, we implement one-step Q-learning as described
in \citet{mnih2016asynchronous}-Alg.~$1$, employing double Q-learning
with $\epsilon$-greedy exploration. We use the same RNN in our policy-based 
approaches to estimate the Q values. A grid search over
exploration rate, exploration rate decay, learning rate, and sync
frequency (between online and target network) is conducted to find the
best hyper-parameters. Unlike our other methods, the Q-learning
baseline uses intermediate rewards, as given by the OpenAI Gym on a
per-step basis. Hence, the Q-learning baseline has a slight advantage
over the policy gradient methods.

In all of the tasks except Copy, our stochastic optimizer uses
mini-batches comprising $400$ policy samples from the model. These
$400$ samples correspond to $40$ different random sequences drawn from
the environment, and $10$ random policy trajectories per sequence. In
other words, we set $K = 10$ and $N = 40$ as defined
in~\eqref{eq:grad2rl} and~\eqref{eq:gradrlrml}.  For \pg, we use the
$10$ samples to subtract the mean of the coefficient of
$\frac{\deriv}{\deriv \btheta}\log\modelp(\bat\mid\bphi)$ which
includes the contribution of the reward and entropy regularization.
For \rml, we use the $10$ trajectories to subtract the mean reward and
normalize the importance sampling weights. We do not subtract the mean
of the normalized importance weights. For the Copy task, we use
mini-batches with $200$ samples using $K=10$ and $N=20$. Experiments
are conducted using Tensorflow~\citep{tensorflow}.
\comment{, which we found to perform better.}

\comment{
For our experiments, we aimed to realize two things: first, if \rml
can give more consistent behavior across different hyper-parameters;
second, if \rml can solve tasks that \pg cannot solve at all.  To this
end, we ran several runs of \rml and \pg on each task.  For each task
we trained a number of models on different values of $\tau$ and
different hyper-parameters.  For \pg, we used $\tau\in\{0, 0.005, 0.01,
0.1\}$, of which we generally found at least one to work well (a
broader search did not yield stronger results for other values of
$\tau$).  For \rml, we used $\tau=0.1$, which we found to work well on
all the tasks.
}

\comment{
\todo{(a broader search did not yield stronger results for other values of
$\tau$)}
}

\comment{
For each training algorithm and value of $\tau$, we ran 60 training
runs, the result of an intersection of three values of learning rate
($\eta\in\{0.1, 0.01, 0.001\}$), four values of clipping value for the
gradient ($c\in\{1, 10, 40, 100\}$), and five random initialization
seeds.  With the exception of Copy, each run was trained using batches
consisting of 10 trajectories each from 40 different resets of the
environment.  Thus, a single batch consisted of 400 total policy
trajectories.  For \pg, we use the 10 trajectories to mean-center the
coefficient of $\frac{\deriv}{\deriv
  \btheta}\log\modelp(\bat\mid\bphi)$.  For \rml, we use the 10
trajectories to both mean-center the rewards and self-normalize the
importance sampling weights.  For the Copy task, we used batches of 10
trajectories each from 20 different resets of the environment, which
we found to perform better.  The runs for Copy, RepeatCopy,
DuplicatedInput, Reverse, ReversedAddition, and BinarySearch were
trained for 2000, 50000, 500, 5000, 50000, and 2000 steps,
respectively.  We found these numbers of steps were mostly enough for
the algorithms to converge for each respective task.

The model architecture we used was an LSTM recurrent neural
network~\citep{lstm} with hidden dimension of 128.  We train using the
Adam optimizer which we chose for its robustness to
hyper-parameters~\citep{adam}.

In addition to these policy gradient methods, we also evaluated the
performance of value-based reinforcement learning.  We implemented
one-step Q-learning as described in \citet{mnih2016asynchronous} and
mentioned in Section \ref{related_work}.  This training algorithm
employs double Q-learning with $\epsilon$-greedy exploration.  We use
the same recurrent neural network in our policy-based methods to act
as the Q-value approximator. We performed a grid search over
exploration rate, exploration rate decay, learning rate, and sync
frequency (between online and target network) to find the best
configuration reported in the results.  Unlike our other methods, we
allow Q-learning to make use of per-step rewards when they are given
(the tasks provided by Gym give per-step rewards), thus giving this
baseline a slight advantage.

\subsection{Experimental details}

We compare \rml to vanilla REINFORCE ($\temp=0$) as well as \pg
($\temp>0$).  Just as for \rml, we only provide an agent with a reward
at the end of an episode (so there is no notion of per-step rewards).

In addition to these policy gradient methods, we also evaluated the
performance of value-based reinforcement learning.  We implemented
one-step Q-learning as described in \citet{mnih2016asynchronous} and
mentioned in Section \ref{related_work}.  This training algorithm
employs double Q-learning with $\epsilon$-greedy exploration.  We use
the same recurrent neural network in our policy-based methods to act
as the Q-value approximator. We performed a grid search over
exploration rate, exploration rate decay, learning rate, and sync
frequency (between online and target network) to find the best
configuration reported in the results.  Unlike our other methods, we
allow Q-learning to make use of per-step rewards when they are given
(the tasks provided by Gym give per-step rewards), thus giving this
baseline a slight advantage.

We find
that not only does \rml solve the considered tasks given the best
hyper-parameters, but also that the \rml-policy gradient is more
robust to the choice of the hyper-parameters, when compared to
the \pg expected reward objective.

\subsection{Baseline methods}
We compare \rml to vanilla REINFORCE ($\temp=0$) as well as \pg ($\temp>0$).  
Just as for \rml, we only provide an agent with a reward at the end of an episode
(so there is no notion of per-step rewards).

In addition to these policy gradient methods, we also evaluated the
performance of value-based reinforcement learning.  We implemented
one-step Q-learning as described in \citet{mnih2016asynchronous} and
mentioned in Section \ref{related_work}.  This training algorithm
employs double Q-learning with $\epsilon$-greedy exploration.  We use
the same recurrent neural network in our policy-based methods to act
as the Q-value approximator. We performed a grid search over
exploration rate, exploration rate decay, learning rate, and sync
frequency (between online and target network) to find the best
configuration reported in the results.  Unlike our other methods, we
allow Q-learning to make use of per-step rewards when they are given
(the tasks provided by Gym give per-step rewards), thus giving this
baseline a slight advantage.

}

\section{Conclusion}
\label{conclusion}

We present a variant of policy gradient, called \rml, which promotes
the exploration of
action sequences that yield rewards larger than what the
model expects. This exploration strategy is the result of importance
sampling from the optimal policy. Our experimental results demonstrate
that \rml significantly outperforms other value and policy based
methods, while being more robust to changes of hyper-parameters.  By
using \rml, we can solve algorithmic tasks like multi-digit addition
from only episodic reward,
which other methods cannot reliably solve even given the best
hyper-parameters. We introduce a new algorithmic task based on binary
search to advocate more research in this area, especially when the
computational complexity of the solution is also of interest. Solving these
tasks is not only important for developing more human-like intelligence in
learning algorithms, but also important for generic
reinforcement learning, where smart and efficient exploration is the
key to successful methods.

\section{Acknowledgment}

We thank Sergey Levine, Irwan Bello, Corey Lynch, George Tucker,
Kelvin Xu, Volodymyr Mnih, and the Google Brain team for insightful
comments and discussions.

\bibliography{iclr.bib}
\bibliographystyle{iclr2017_conference}

\newpage
\appendix

\ifdefined\DRAFT

\section{Analysis of the choice of temperature}
\label{temp_analysis}
When choosing $\temp$ we want to choose it to be small
enough so that the gradient of our objective encourages
our model to go towards high-reward regions rather than 
explore in low-reward regions.  For simplicity, assume
that our environment is a one-action environment with 
action space $\A$ and that the maximal reward is $1$ attained from action $\bat_{\text{hi}}$
and the minimal reward is $0$ attained from action $\bat_{\text{lo}}$.  
Moreover, assume that this environment exhibits sparse reward - most of the actions
yield little or no reward.

A good $\temp$ that would help our model find the best action would
have the coefficient of $\frac{\deriv}{\deriv \btheta}\log \modelp(\bast \mid \bphii)$
in the gradient given by \eqref{eq:grad2rl} larger at $\bat_{\text{hi}}$ than 
at $\bat_{\text{lo}}$.  That is, we want
\begin{equation}
\reward{\bat_{\text{hi}}}{\bphii} - \temp \log \modelp(\bat_{\text{hi}} \mid \bphii) >
\gamma\left(\reward{\bat_{\text{lo}}}{\bphii} - \temp \log \modelp(\bat_{\text{lo}} \mid \bphii)\right),
\end{equation}
for some large $\gamma$.  Assuming we are in a good region of policy space already,
we can take $\modelp(\bat_{\text{hi}}\mid\bphii)$ close to 1 and 
$\modelp(\bat_{\text{lo}} \mid \bphii)\le\frac{1}{|\A|}$.
Thus, we should choose $\temp$ so that
\begin{equation}
\temp < \frac{1}{\gamma \log|\A|}.
\end{equation}

For \rml the same logic would also look at the coefficient of $\frac{\deriv}{\deriv \btheta}\log \modelp(\bast \mid \bphii)$
in the gradient.  We consider the unnormalized importance weights, so that the coefficient is

\begin{equation}
\reward{\bast}{\bphii} + \frac{\temp}{\modelp(\bast \mid \bphii)}\exp\left(\reward{\bast}{\bphii} / \temp\right) / Z(\bphi).
\end{equation}

Given our assumption that our action space yields sparse rewards with max 1,
we approximate $Z(\bphi)=\sum_{\bat\in \A}\exp\left(\reward{\bat}{\bphii} / \temp\right)$ as 
$\exp\left(\epsilon+1/\temp\right)$, where $\epsilon$ is small.

Thus, the same logic we used for \pg encourages us to choose $\temp$ for \rml 
so that

\begin{multline}
\reward{\bat_{\text{hi}}}{\bphii} + 
\frac{\temp}{\modelp(\bat_{\text{hi}} \mid \bphii)}\exp\left(\reward{\bat_{\text{hi}}}{\bphii} / \temp\right) / Z(\bphi) > \\
\gamma \left( 
\reward{\bat_{\text{lo}}}{\bphii} +                                                                                       
\frac{\temp}{\modelp(\bat_{\text{lo}} \mid \bphii)}\exp\left(\reward{\bat_{\text{lo}}}{\bphii} / \temp\right) / Z(\bphi) \right).
\end{multline}

Thus,
\begin{equation}
1 + 
\temp\exp\left(\frac{- \epsilon}{\temp} \right) >
\gamma \left(                                                                                   
\temp |\A| \exp\left(\frac{-1-\epsilon}{\temp} \right)\right).
\end{equation}

It is enough for $\temp$ to be small enough so that
\begin{equation}
\temp < \frac{1}{\log\gamma + \log|\A|}.
\end{equation}

So we conclude that \pg necessitates a much smaller choice of temperature
than \rml.

\else{} 
\fi 

\section{Optimal Policy for the \rml Objective}
\label{sec:appxB}
To derive the form of the optimal policy for the
\rml objective \eqref{eq:objrml2},
note that for each $\bphi$ one would like to maximize
\begin{equation}
\sum_{\bat\in\A}\Big[
\modelp(\bat)\,r(\bat)
+\temp\,\optp(\bat) \log \modelp(\bat)
\Big]
~,
\end{equation}
subject to the constraint $\sum_{\bat\in\A}\modelp(\bat)=1$.
To enforce the constraint, we introduce a Lagrange multiplier $\alpha$
and aim to maximize
\begin{equation}
\sum_{\bat\in\A}\Big[
\modelp(\bat)\,r(\bat)
+\temp\,\optp(\bat) \log \modelp(\bat)
-\alpha\modelp(\bat)
\Big]+\alpha
~.
\label{eq:lag}
\end{equation}
Since the gradient of the Lagrangian \eqref{eq:lag} 
with respect to $\btheta$
is given by
\begin{equation}
\sum_{\bat\in\A}
\frac{\deriv\modelp(\bat)}{\deriv\btheta}
\Big[
r(\bat)
+\temp\,\frac{\optp(\bat)}{\modelp(\bat)}
-\alpha
\Big]
~,
\end{equation}
the optimal choice for $\modelp$
is achieved by setting
\begin{equation}
\modelp(\bat)=
\frac{\temp\,\optp(\bat)}{\alpha-r(\bat)}
~~~\mbox{for all}~~~\bat\in\A~,
\end{equation}
forcing the gradient to be zero.
The Lagrange multiplier $\alpha$ can then be chosen 
so that 
$\sum_{\bat\in\A}\modelp(\bat)=1$
while also satisfying
$\alpha>\max_{\bat\in\A}r(\bat)$;
see {\em e.g.}\ \citep{golub73}.

\comment{
We can express the \rml objective given in \eqref{eq:objrml2} as
\begin{equation}
\modelp^{T}\log\optp + \optpt\log\modelp.
\end{equation}
Our aim is to maximize this objective under the constraint
$\modelp^{T}\vec{1} = 1$.  
Note, that the additional constraint $\modelp\ge\vec{0}$ can be enforced by
letting $\modelp=\exp(\modelp^\prime)$, which does not change our analysis.
We can use the method of Lagrange
Multipliers to optimize the constrained objective.  We introduce an unconstrained
real $\alpha$ to our objective:
\begin{equation}
\modelp^{T}\log\optp + \optpt\log\modelp + \alpha (\modelp^{T}\vec{1} - 1).
\end{equation}
The optimum of this objective occurs when the gradient is 0.  Setting the 
gradient with respect to $\alpha$ to 0 yields $\modelp^{T}\vec{1} = 0$.
Setting the gradient with respect to $\modelp$ to 0 yields
\begin{equation}
\modelp^\prime \cdot \log\optp + \optp \cdot \frac{\modelp^\prime}{\modelp} + \alpha\modelp^\prime = 0.
\end{equation}
Factoring out $\modelp^\prime$, we find
\begin{equation}
\log\optp + \frac{\optp}{\modelp} + \alpha = 0,
\end{equation}
which yields
\begin{equation}
\modelp = \frac{-\optp}{\alpha + \log\optp}.
\end{equation}

Therefore, the optimal $\modelp$ is given by $\exp\left(r / \temp - \log(-\alpha - r / \temp)\right)$,
where $\alpha$ is chosen so that $\modelp$ forms a valid probability distribution.
}

\section{Robustness to Hyper-parameters}
\label{appendix}
Tables \ref{results-copy}--\ref{results-bin} provide more details on
different cells of \tabref{gym-results}.  Each table presents the
results of \pg using the best temperature $\temp$ \vs \rml
with $\temp = 0.1$ on a variety of learning rates and clipping
values. Each cell is the number of trials out of $5$ random restarts
that succeed at solving the task using a specific $\eta$ and $c$.

\begin{table}[H]
\caption{Copy -- number of successful attempts out of $5$.}
\label{results-copy}
\begin{center}
\begin{tabular}{c|c|c|c||c|c|c|}
\cline{2-7}
 & \multicolumn{3}{|c||}{\bold\pg ~~~($\temp=0.01$)} & \multicolumn{3}{c|}{\bold\rml ~~~($\temp=0.1$)} \\ \cline{2-7}
 & {\small$\eta=0.1$} &{\small$\eta=0.01$} & {\small$\eta=0.001$}
 & {\small$\eta=0.1$} &{\small$\eta=0.01$} & {\small$\eta=0.001$} \\ \hline

\multicolumn{1}{|c|}{\small$c=1$} & 3 & 5 & 5 & 5 & 5 & 2 \\ \cline{1-7}
\multicolumn{1}{|c|}{\small$c=10$} & 5 & 4 & 5 & 5 & 5 & 3 \\ \cline{1-7}
\multicolumn{1}{|c|}{\small$c=40$} & 3 & 5 & 5 & 4 & 4 & 1 \\ \cline{1-7}
\multicolumn{1}{|c|}{\small$c=100$} & 4 & 5 & 5 & 4 & 5 & 2 \\ \cline{1-7}

\end{tabular}
\end{center}
\vspace*{-.5cm}
\end{table}

\begin{table}[H]
\caption{DuplicatedInput -- number of successful attempts out of $5$.}
\label{results-dup}
\begin{center}
\begin{tabular}{c|c|c|c||c|c|c|}
\cline{2-7}
 & \multicolumn{3}{|c||}{\bold\pg ~~~($\temp=0.01$)} & \multicolumn{3}{c|}{\bold\rml ~~~($\temp=0.1$)} \\ \cline{2-7}
 & {\small$\eta=0.1$} &{\small$\eta=0.01$} & {\small$\eta=0.001$}
 & {\small$\eta=0.1$} &{\small$\eta=0.01$} & {\small$\eta=0.001$} \\ \hline

\multicolumn{1}{|c|}{\small$c=1$} & 3 & 5 & 3 & 5 & 5 & 5 \\ \cline{1-7}
\multicolumn{1}{|c|}{\small$c=10$} & 2 & 5 & 3 & 5 & 5 & 5 \\ \cline{1-7}
\multicolumn{1}{|c|}{\small$c=40$} & 4 & 5 & 3 & 5 & 5 & 5 \\ \cline{1-7}
\multicolumn{1}{|c|}{\small$c=100$} & 2 & 5 & 4 & 5 & 5 & 5 \\ \cline{1-7}

\end{tabular}
\end{center}
\vspace*{-.5cm}
\end{table}

\begin{table}[H]
\caption{RepeatCopy -- number of successful attempts out of $5$.}
\label{results-repeat}
\begin{center}
\begin{tabular}{c|c|c|c||c|c|c|}
\cline{2-7}
 & \multicolumn{3}{|c||}{\bold\pg ~~~($\temp=0.01$)} & \multicolumn{3}{c|}{\bold\rml ~~~($\temp=0.1$)} \\ \cline{2-7}
 & {\small$\eta=0.1$} &{\small$\eta=0.01$} & {\small$\eta=0.001$}
 & {\small$\eta=0.1$} &{\small$\eta=0.01$} & {\small$\eta=0.001$} \\ \hline

\multicolumn{1}{|c|}{\small$c=1$} & 0 & 1 & 0 & 0 & 2 & 0 \\ \cline{1-7}
\multicolumn{1}{|c|}{\small$c=10$} & 0 & 0 & 2 & 0 & 4 & 0 \\ \cline{1-7}
\multicolumn{1}{|c|}{\small$c=40$} & 0 & 0 & 1 & 0 & 2 & 0 \\ \cline{1-7}
\multicolumn{1}{|c|}{\small$c=100$} & 0 & 0 & 3 & 0 & 3 & 0 \\ \cline{1-7}

\end{tabular}
\end{center}
\vspace*{-.5cm}
\end{table}

\begin{table}[H]
\caption{Reverse -- number of successful attempts out of $5$.}
\label{results-reverse}
\begin{center}
\begin{tabular}{c|c|c|c||c|c|c|}
\cline{2-7}
 & \multicolumn{3}{|c||}{\bold\pg ~~ ($\temp=0.1$)} & \multicolumn{3}{c|}{\bold\rml ~~~($\temp=0.1$)} \\ \cline{2-7}
 & {\small$\eta=0.1$} &{\small$\eta=0.01$} & {\small$\eta=0.001$}
 & {\small$\eta=0.1$} &{\small$\eta=0.01$} & {\small$\eta=0.001$} \\ \hline

\multicolumn{1}{|c|}{\small$c=1$} & 1 & 1 & 0 & 0 & 0 & 0 \\ \cline{1-7}
\multicolumn{1}{|c|}{\small$c=10$} & 0 & 1 & 0 & 0 & 4 & 0 \\ \cline{1-7}
\multicolumn{1}{|c|}{\small$c=40$} & 0 & 2 & 0 & 0 & 2 & 1 \\ \cline{1-7}
\multicolumn{1}{|c|}{\small$c=100$} & 1 & 0 & 0 & 0 & 2 & 1 \\ \cline{1-7}

\end{tabular}
\end{center}
\end{table}

\begin{table}[H]
\caption{ReversedAddition -- number of successful attempts out of $5$.}
\label{results-revaddition}
\begin{center}
\begin{tabular}{c|c|c|c||c|c|c|}
\cline{2-7}
 & \multicolumn{3}{|c||}{\bold\pg ~~~($\temp=0.01$)} & \multicolumn{3}{c|}{\bold\rml ~~~($\temp=0.1$)} \\ \cline{2-7}
 & {\small$\eta=0.1$} &{\small$\eta=0.01$} & {\small$\eta=0.001$}
 & {\small$\eta=0.1$} &{\small$\eta=0.01$} & {\small$\eta=0.001$} \\ \hline

\multicolumn{1}{|c|}{\small$c=1$} & 0 & 0 & 0 & 0 & 0 & 4 \\ \cline{1-7}
\multicolumn{1}{|c|}{\small$c=10$} & 0 & 0 & 0 & 0 & 3 & 2 \\ \cline{1-7}
\multicolumn{1}{|c|}{\small$c=40$} & 0 & 0 & 0 & 0 & 0 & 5 \\ \cline{1-7}
\multicolumn{1}{|c|}{\small$c=100$} & 0 & 0 & 1 & 0 & 1 & 3 \\ \cline{1-7}

\end{tabular}
\end{center}
\end{table}

\begin{table}[H]
\caption{BinarySearch -- number of successful attempts out of $5$.}
\label{results-bin}
\begin{center}
\begin{tabular}{c|c|c|c||c|c|c|}
\cline{2-7}
 & \multicolumn{3}{|c||}{\bold\pg ~~~($\temp=0.01$)} & \multicolumn{3}{c|}{\bold\rml ~~~($\temp=0.1$)} \\ \cline{2-7}
 & {\small$\eta=0.1$} &{\small$\eta=0.01$} & {\small$\eta=0.001$}
 & {\small$\eta=0.1$} &{\small$\eta=0.01$} & {\small$\eta=0.001$} \\ \hline

\multicolumn{1}{|c|}{\small$c=1$} & 0 & 0 & 0 & 0 & 4 & 0 \\ \cline{1-7}
\multicolumn{1}{|c|}{\small$c=10$} & 0 & 1 & 0 & 0 & 3 & 0 \\ \cline{1-7}
\multicolumn{1}{|c|}{\small$c=40$} & 0 & 0 & 0 & 0 & 3 & 0 \\ \cline{1-7}
\multicolumn{1}{|c|}{\small$c=100$} & 0 & 0 & 0 & 0 & 2 & 0 \\ \cline{1-7}

\end{tabular}
\end{center}
\end{table}

\section{Generalization to Longer Sequences}
\label{appendix-generalization}

Table \ref{gen-results} provides a more detailed look into the generalization performance
of the trained models on Copy, DuplicatedInput, and ReversedAddition.  The tables show
how the number of models which can solve the task correctly drops off as the length 
of the input increases.

\begin{table}[!ht]
\caption{Generalization results.
Each cell includes the number of runs out of $60$ different
hyper-parameters and random initializations that achieve $100\%$
accuracy on input of length up to the specified length.  The bottom
row is the maximal length ($\le\!2000$) up to which at least one model
achieves $100\%$ accuracy.}
\label{gen-results}
\begin{center}
\begin{tabular}{c|c|c||c|c||c|c|}
\cline{2-7}
 & \multicolumn{2}{c||}{\bold Copy} & \multicolumn{2}{c||}{\bold DuplicatedInput}
 & \multicolumn{2}{c|}{\bold ReversedAddition} \\ \cline{2-7}
 & \multicolumn{1}{c|}{\bold \pg} & \multicolumn{1}{c||}{\bold \rml}
 & \multicolumn{1}{c|}{\bold \pg} & \multicolumn{1}{c||}{\bold \rml}
 & \multicolumn{1}{c|}{\bold \pg} & \multicolumn{1}{c|}{\bold \rml} \\ \hline
\multicolumn{1}{|c|}{\bold 30}	& 54 & 45 & 44 & 60 & 1 & 18 \\ \hline
\multicolumn{1}{|c|}{\bold 100}	& 51 & 45 & 36 & 56 & 0 & 6 \\ \hline
\multicolumn{1}{|c|}{\bold 500}	& 27 & 22 & 19 & 25 & 0 & 5 \\ \hline
\multicolumn{1}{|c|}{\bold 1000}	& 3 & 2 & 12 & 17 & 0 & 5 \\ \hline
\multicolumn{1}{|c|}{\bold 2000}	& 0 & 0 & 6 & 9 & 0 & 5 \\ \hline
\multicolumn{1}{|c|}{\bold Max}	& 1126 & 1326 & 2000 & 2000 & 38 & 2000 \\ \hline
\end{tabular}
\end{center}
\end{table}

\section{Example Execution Traces}
\label{appendix-traces}
\begin{figure}[H]
\begin{center}
\includegraphics[width=0.9\textwidth]{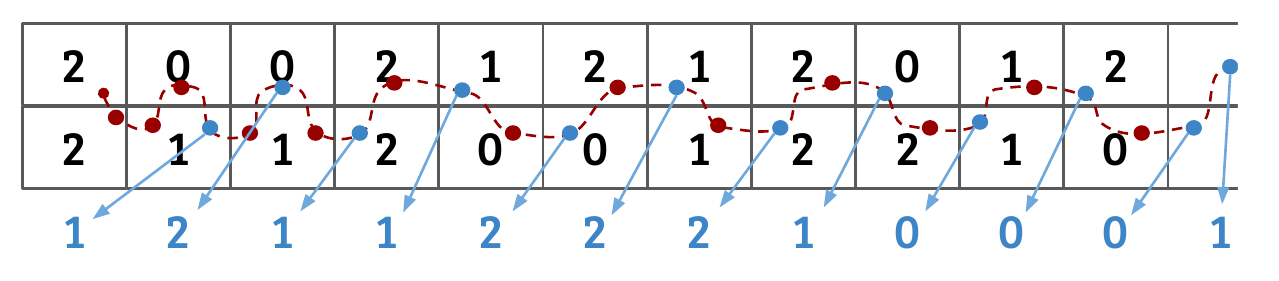}
\end{center}
\caption{A graphical representation of a trained addition agent.  
The agent begins at the top left corner of a $2\times n$ grid of
ternary digits.  At each time step, it may move to the left, right, up,
or down (observing one digit at a time) and optionally write to output.}
\label{fig:addtrace}
\end{figure}

We provide the traces of two trained agents on the ReversedAddition
task (\figref{fig:addtrace})
and the BinarySearch task (\tabref{binarysearch-trace}).

\begin{table}[H]
\caption{Example trace on the BinarySearch task where $n=512$ and the number
to find is at position $100$.  At time $t$ the agent observes $\bs_t$ from the environment
and samples an action $\ba_t$.  We also include the inferred range of indices to which the
agent has narrowed down the position of $x$.  We see that the first several steps of the agent
follow a binary search algorithm.  However, at some point the agent switches to a linear search}
\label{binarysearch-trace}
\begin{center}
\begin{tabular}{|ccc|c|c|c|} \hline
$R_0$ & $R_1$ & $R_2$ & $\bs_t$ & \multicolumn{1}{c|}{$\ba_t$} & Inferred range \\ \hline
512 & 0 & 0 & -- & $\mathrm{AVG}(2)$ & $[0, 512)$ \\ \hline
512 & 0 & 256 & -- & $\mathrm{CMP}(2)$ & $[0, 512)$ \\ \hline
512 & 0 & 256 & $<$ & $\mathrm{DIV}(0)$ & $[0, 256)$ \\ \hline
256 & 0 & 256 & -- & $\mathrm{AVG}(2)$ & $[0, 256)$ \\ \hline
256 & 0 & 128 & -- & $\mathrm{CMP}(2)$ & $[0, 256)$ \\ \hline
256 & 0 & 128 & $<$ & $\mathrm{DIV}(0)$ & $[0, 128)$ \\ \hline
128 & 0 & 128 & -- & $\mathrm{AVG}(2)$ & $[0, 128)$ \\ \hline
128 & 0 & 64 & -- & $\mathrm{CMP}(2)$ & $[0, 128)$ \\ \hline
128 & 0 & 64 & $>$ & $\mathrm{AVG}(1)$ & $(64, 128)$ \\ \hline
128 & 96 & 64 & -- & $\mathrm{CMP}(1)$ & $(64, 128)$ \\ \hline
128 & 96 & 64 & $>$ & $\mathrm{AVG}(2)$ & $(96, 128)$ \\ \hline
128 & 96 & 112 & -- & $\mathrm{CMP}(2)$ & $(96, 128)$ \\ \hline
128 & 96 & 112 & $<$ & $\mathrm{AVG}(1)$ & $(96, 112)$ \\ \hline
128 & 120 & 112 & -- & $\mathrm{CMP}(2)$ & $(96, 112)$ \\ \hline
128 & 120 & 112 & $<$ & $\mathrm{DIV}(1)$ & $(96, 112)$ \\ \hline
128 & 60 & 112 & -- & $\mathrm{AVG}(2)$ & $(96, 112)$ \\ \hline
128 & 60 & 94 & -- & $\mathrm{CMP}(2)$ & $(96, 112)$ \\ \hline
128 & 60 & 94 & $>$ & $\mathrm{AVG}(1)$ & $(96, 112)$ \\ \hline
128 & 111 & 94 & -- & $\mathrm{CMP}(1)$ & $(96, 112)$ \\ \hline
128 & 111 & 94 & $<$ & $\mathrm{INC}(1)$ & $(96, 111)$ \\ \hline
128 & 112 & 94 & -- & $\mathrm{INC}(2)$ & $(96, 111)$ \\ \hline
128 & 112 & 95 & -- & $\mathrm{CMP}(2)$ & $(96, 111)$ \\ \hline
128 & 112 & 95 & $>$ & $\mathrm{INC}(2)$ & $(96, 111)$ \\ \hline
128 & 112 & 96 & -- & $\mathrm{CMP}(2)$ & $(96, 111)$ \\ \hline
128 & 112 & 96 & $>$ & $\mathrm{INC}(2)$ & $(96, 111)$ \\ \hline
128 & 112 & 97 & -- & $\mathrm{CMP}(2)$ & $(96, 111)$ \\ \hline
128 & 112 & 97 & $>$ & $\mathrm{INC}(2)$ & $(97, 111)$ \\ \hline
128 & 112 & 98 & -- & $\mathrm{CMP}(2)$ & $(97, 111)$ \\ \hline
128 & 112 & 98 & $>$ & $\mathrm{INC}(2)$ & $(98, 111)$ \\ \hline
128 & 112 & 99 & -- & $\mathrm{CMP}(2)$ & $(98, 111)$ \\ \hline
128 & 112 & 99 & $>$ & $\mathrm{INC}(2)$ & $(99, 111)$ \\ \hline
128 & 112 & 100 & -- & $\mathrm{CMP}(2)$ & $(99, 111)$ \\ \hline
128 & 112 & 100 & $=$ & -- & -- \\ \hline
\end{tabular}
\end{center}
\end{table}

\section{Variance of Importance Weights}
\label{appendix-variance}

\begin{figure}[H]
\begin{center}
\includegraphics[width=1.0\textwidth]{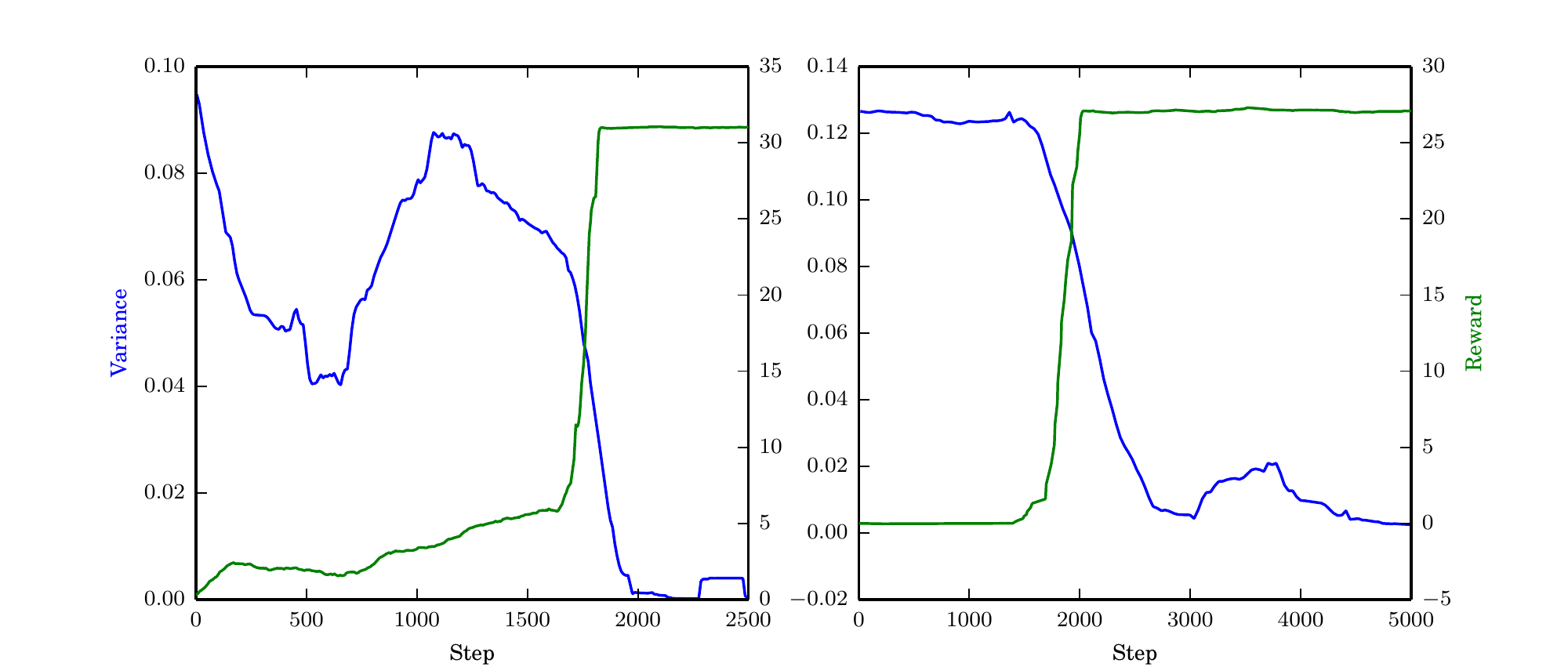}
\end{center}
\caption{This plot shows the variance of the importance weights in the \rml updates
as well as the average reward for two successful runs.  
We see that the variance starts off high and reaches
near zero towards the end when the optimal policy is found.  In the first plot, 
we see a dip
and rise in the variance which corresponds to a plateau and then increase
in the average reward.}
\end{figure}

\section{A Simple Bandit Task}
\label{bandit_task}

\begin{figure}[H]
\begin{center}
\includegraphics[width=0.6\textwidth]{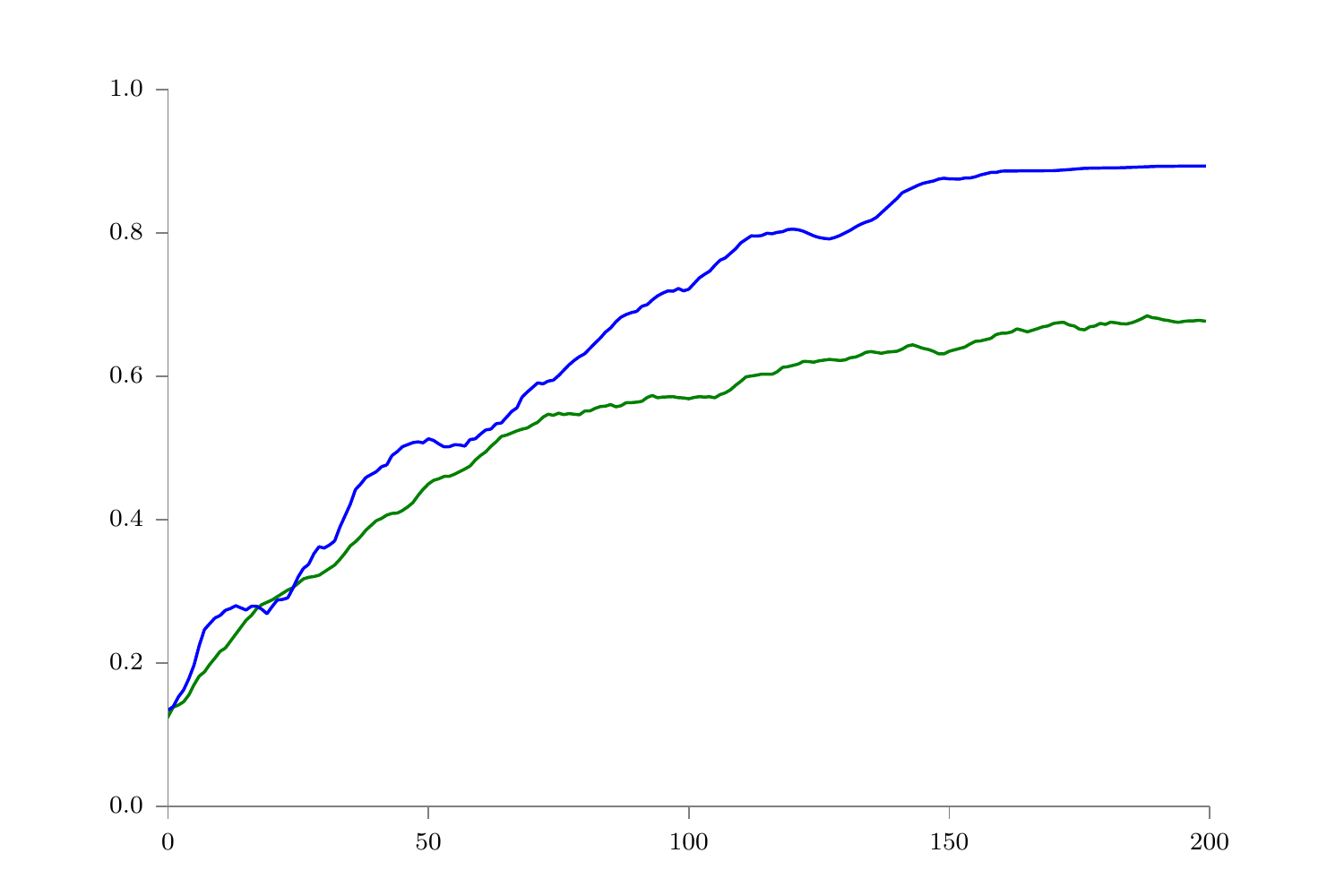}
\end{center}
\caption{In this plot we present the average performance of \rml (blue)
and \pg (green) over 100 repeats of a bandit-like task 
after choosing optimal hyperparameters for each method.
In the task, the agent chooses one of 10,000 actions at each step
and receives a payoff corresponding to the entry in a reward vector 
$\vec{r} = (r_1,...,r_{10,000})$ such that $r_i=u_i^\beta$,
where $u_i\in[0,1)$ has been sampled randomly and independently 
from a uniform distribution.
We parameterize the policy with a weight vector $\theta\in \mathbb{R}^{30}$
such that $\pi_\theta(a)\propto\exp(\vec{\phi}^{(a)}\cdot\theta)$,
where the basis vectors $\vec{\phi}^{(a)}\in\mathbb{R}^{30}$
for each action are sampled from a standard normal distribution.
The plot shows the average rewards obtained by setting $\beta=8$
over $100$ experiments, consisting of
$10$ repeats
(where $\vec{r}$ and $\vec{\Phi}=(\vec{\phi}^{(1)},\dots,\vec{\phi}^{(10,000)})$ are redrawn at the start of each repeat),
with 10 random restarts within each repeat
(keeping $\vec{r}$ and $\vec{\Phi}$ fixed but reinitializing $\theta$).
Thus, this task presents a relatively simple problem with a large
action space, and we again see that \rml outperforms \pg.
}
\end{figure}

\end{document}